\documentclass[conference,8pt]{IEEEtran}
\IEEEoverridecommandlockouts
\usepackage{times}  % DO NOT CHANGE THIS
\usepackage{helvet} % DO NOT CHANGE THIS
\usepackage{courier}  % DO NOT CHANGE THIS
\usepackage[hyphens]{url}  % DO NOT CHANGE THIS
\usepackage{graphicx} % DO NOT CHANGE THIS
\usepackage{graphicx}  % DO NOT CHANGE THIS
\usepackage{float}
\usepackage{multirow}
\usepackage{caption}
\usepackage{algorithm, algorithmic}
\usepackage{amsmath}
\usepackage{amsfonts}
\usepackage{booktabs}
\newcommand{\RNum}[1]{\uppercase\expandafter{\romannumeral #1\relax}}

\usepackage{subfig}

\usepackage{booktabs}
\usepackage{multirow}

\usepackage{color,soul}

\newtheorem{theorem}{Theorem}[section]

\newtheorem{lemma}[theorem]{Lemma}

\title{A Geometric Approach to Online Streaming Feature Selection} %\\ Supplementary Materials}
\author{Salimeh~Yasaei~Sekeh\thanks{S. Yasaei Sekeh is with School of Computing and Information Science at the University of Maine, Orono}, Madan~Ravi~Ganesh$^*$, Shurjo~Banerjee$^*$, Jason~J.~Corso, and Alfred~O.~Hero\thanks{M. Ravi Ganesh, S. Banerjee,J.J, Corso, A.O. Hero are with Electrical Engineering and Computer Science Department at the University of Michigan}\thanks{$^*$Both authors contributed equally to this research.}
}

\begin{document}
\maketitle

\begin{abstract}
Online Streaming Feature Selection (OSFS) is a sequential learning problem where individual features across all samples are made available to algorithms in a streaming fashion.
While the OSFS problem setup is popular, it suffers from limitations ranging from weak assumptions in its problem setup, inefficiencies in its best algorithms, and the lack of a consistent methodology to compare algorithms. 
In this work, we systematically address these three limitations.
Firstly, we assert that OSFS's main assumption of having data from all the samples available at runtime is unrealistic and introduce a new setting where features and samples are streamed concurrently called OSFS with Streaming Samples (OSFS-SS). \
Secondly, the primary OSFS method, SAOLA \cite{Yuetal2014} utilizes an unbounded mutual information measure and requires multiple comparison steps between the stored and incoming feature sets to evaluate a feature's importance. 
We introduce Geometric Online Adaption, an algorithm that requires relatively less feature comparison steps and uses a bounded conditional geometric dependency measure. 
Our algorithm outperforms several OSFS baselines including SAOLA on a variety of datasets.
We also extend SAOLA to work in the OSFS-SS setting and show that GOA continues to achieve the best results.
Thirdly, the current paradigm of the OSFS algorithm comparison is flawed. 
Algorithms are measured by comparing the number of features used and the accuracy obtained by the learner, two properties that are fundamentally at odds with one another.
Without fixing a limit on either of these properties, the qualities of features obtained by different algorithms are incomparable.
We try to rectify this inconsistency by fixing the maximum number of features available to the learner and comparing algorithms in terms of their accuracy. 
Additionally, we characterize the behaviour of SAOLA and GOA on feature sets derived from popular deep convolutional featurizers.
\end{abstract}

\section{Introduction}
\label{sec:introduction}
%Let us motivate the field of Feature Selection (FS) with a simple example. Consider a doctor diagnosing a patient with disease; given the patient's health history and test results, the doctor will focus on relevant pieces of information to make her final diagnosis. This usage of a subset of the information available in the decision making process inspires the field of FS~\cite{tibshirani1996regression,kohavi1997wrappers,dash2003consistency,yu2003feature}. By limiting the number of features required to perform a task, FS allows for the training of faster models that utilize lower computational and storage requirements while ideally achieving similar performance to working with all the features. 

%Let us motivate the field of Online Streaming Feature Selection with a simple example. Consider a driver learning how to travel to a new destination using a map. The driver is likely to trace out his path from his origin to the final destination while ignoring much of the information (while  mapping system to drive to a new destination. The driver must make navigational decisions based on information provided to them by system. While they are able to query the device for additional information pertaining to their surroundings, they are more likely to base most decisions of information as it is made available by the system. Such real world 

Online Streaming Feature Selection (OSFS)~\cite{Zhouetal2005,wu2010online} is a sequential learning problem where individual features across all samples are made available to algorithms in streaming fashion. Algorithms are allowed a single pass at features and are tasked with maintaining a fixed size decision-relevant list of them by disregarding uninformative old ones (i.e. measuring feature redundancy) and incorporating informative new ones (i.e. measuring feature relevancy). While OSFS is an important direction of research due to its applicability to low-memory usage systems, high-throughput systems, and other scenarios, it suffers from limitations  which include weak assumptions in its problem statement, inefficiencies in its most popular methods, and the lack of a consistent methodology to compare algorithms. In this work, we propose a novel setup and technique in which we systematically address these limitations.

The primary limitation of OSFS lies in the assumption that data from all the samples must be available at runtime. 
We assert that this unrealistic and introduce a new setting where both features and samples are streamed concurrently. 
We call this new setting Online Streaming Feature Selection with Streaming Samples (OSFS-SS), where we experiment with multiple sizes of the samples streamed to learners (Section \ref{subsec:osfsss}).

For the second limitation, many methods have been proposed for the OSFS learning problem. One of the earliest works, \cite{Zhouetal2005}, introduced the Alpha-Investing algorithm. Here new features are added to the minimal subset via dynamic thresholding of a p-statistic between current features and incoming features. Alpha-investing, however, does not attempt to limit the number of features that are kept in the current feature set.
While \cite{wu2010online,Wuetal.2013} were the first to introduce concepts of feature relevance and redundancy to prune and maintain small subsets of features for this task, SAOLA (Scalable and Accurate Online feature selection) ~\cite{Yuetal2014}, the most popular approach, took it further by using mutual information (MI)~\cite{Cover1991ElementsOI} to measure intra-feature redundancy and dependency. 
However, SAOLA uses multiple comparison steps in their algorithm to evaluate the importance of features, which makes it inefficient, as well as the histogram estimator which is not bounded.
Hence, their hyper-parameter tuning for thresholds is not restricted to small search space.

We address the limitations of these methods by introducing the Geometric Online Adaption (GOA) algorithm. 
Our algorithm uses the geometry of samples and their relative spacing to construct the comparisons that help retain or remove features.
This ensures only two comparison steps in GOA.
We also use a class-conditional geometric dependency measure (CGD) ~\cite{ICASSPsalimeh2019} to calculate intra-feature relevancy and redundancy. 
CGD is useful as it is bounded allowing us to fine-tune our hyper-parameters within a truly restricted space.
When compared to SAOLA, our method contains a lower number of comparison steps while ensuring true bounds for the thresholds. 
Under GOA's assumptions, we show that there exists a feature in the current optimal feature subset such that if the new incoming feature is irrelevant to the class label given that feature, then both the given and incoming feature are dependent (Theorem \ref{thm.1}). We validate our method against the algorithms described above across multiple datasets (Section \ref{osfs-fixed-sample}).

Apart from evaluating GOA on OSFS, we augment SAOLA to work in the OSFS-SS setting as a baseline.
We use this to showcase how GOA continues to achieve the best results. 
The new streaming setting also allows us to apply SAOLA and GOA to features derived from publicly available convolutional neural network (CNN) featurizers. 
These experiments help us test the presence of redundancy in these deep featurizers and evaluate how they are different from the human-annotated feature sets that normally dominate this line of work.

The third limitation is the lack of a consistent methodology to compare OSFS algorithms. 
Generally, they are compared by tabulating the size of the feature sets used and the accuracy of the learners. However, these are two properties that are at odds with each other - the more the number of features used, the higher the accuracy and vice versa. 
As such, it is difficult to decide when one algorithm is truly outperforming another. 
In this work, we compare our algorithm to every baseline individually by fixing the number of features used by GOA to be less than or equal to that of the baseline being compared against.
Under this setting, an algorithm's performance can then be compared using the learners' accuracy.
We believe that while our approach is simple, it rectifies the problem.

\section{Related Work}
\label{sec:related_work}
In our related work, we first outline prior OSFS methods before describing experimental setups similar in scope but ultimately different from our OSFS-SS extension.
%Finally we review 
%First we describe the graph-based geometric dependency measure used. Second we describe relevant work from the OSFS literature. Finally we talk about works that are similar in scope but ultimate different in the final objective to our OSFS-SS extensions. 

%\subsection{CGD Measure}
% In the multi-class classification context, the CGD measure is used at the core of our proposed GOA method to determine relevancy between features. 
% The use of similar graph-based measures and its advantages in multi-class classification problems have been studied in \cite{SalimehetalAlertton2018} and \cite{ICASSPsalimeh2019} in non-streaming settings.
% Further, in \cite{ICASSPsalimeh2019} it has been shown that in the multi-class classification setting the conditional geometric dependency measures \cite{SalimehEntropy2018} between features, given class label variables as a relevancy measure outperforms several feature selection baselines. As such we believe the usage of this measure in GOA's algorithm makes sense and is key to its success. 

\subsection{Online Streaming Feature Selection}

 Among works that span OSFS, Alpha-investing~\cite{Zhouetal2005} forms one of the earliest methods which deals with streaming feature selection problems. In the work, the authors dynamically adjust a threshold on the $p$-statistic for new features to enter the model in order to control the false discovery rate~\cite{benjamini1995controlling}. While their strategy is effective, no additional steps are taken to reduce the number of features collected.
 
 Online Streaming Feature Selection~\cite{wu2010online} (from whence we derive our terminology) and its faster adaptation, Fast-OSFS~\cite{Wuetal.2013}, were developed to use probabilistic concepts of Markov Blankets to measure intra-feature relevancy and redundancy. Both methods break up the redundancy analysis into two steps,  inner-redundancy (the redundancy between current features and incoming features) and outer-redundancy (redundancy between incoming features and features collected in the minimal subset thus far). The algorithm proposed in this study was evaluated on a large scale using several datasets, many of which are used in our work.
 
 A more recent approach, SAOLA~\cite{Yuetal2014}, is derived from a lower bound of correlations between features using pairwise comparisons. The method uses histogram-based mutual information (MI) to perform comparisons for inner-redundancy and outer-redundancy operations and maintains a parsimonious set of features for training over a longer duration. 
 
 A new algorithm called OSNRRSAR-SA~\cite{eskandari2016}  was proposed to resolve online streaming feature selection from the rough sets (RS) perspective ~\cite{roughset}. This algorithm adopts a Rough sets-based approach on feature significance to remove non-relevant features. 
 A survey of feature selection approaches with more detailed descriptions of each method is provided in \cite{alnuaimi2019streaming} and \cite{tang2014feature}. 

\subsection{Online Streaming Feature Selection with Streaming Samples}
 To the best of our knowledge, we are the first to propose experiments in the OSFS-SS domain though there are works that share key characteristics.  
 An important goal of feature selection from streaming samples is to approximate missing data distributions using a subset of the selected features~\cite{li2016spatially,liu2019adaptive}.
 Taking it a step further, unsupervised methods use low-rank approximations as a matrix completion mechanism to improve and match performance~\cite{huang2015unsupervised}.
 \cite{shao2016online} used a clustering alternative to simple low-rank approximations for unsupervised feature selection in streaming multi-view data.
 
 The key difference between our proposed OSFS-SS experiments and the above methods is they can take multiple passes over the data and assume that all features are available throughout the training process while OSFS-SS allows only a single pass over all features and samples.

%\newcounter{MYtempeqncnt}

% \renewcommand{\thefootnote}{}
% \newtheorem{theorem}{Theorem}[section]
% \newtheorem{acknowledgment}[theorem]{Acknowledgment}
% \newtheorem{corollary}[theorem]{Corollary}
% \newtheorem{definition}[theorem]{Definition}
% \newtheorem{example}[theorem]{Example}
% \newtheorem{lemma}[theorem]{Lemma}
% \newtheorem{proposition}[theorem]{Proposition}
% \newtheorem{remark}[theorem]{Remark}
% \newtheorem{conjecture}[theorem]{Conjecture}

%\renewcommand{\encodingdefault}{T1}

%\newtheorem*{thm}{Theorem}

\def\cmp{{\complement}}

\def\tA{{\tt A}}
\def\tB{{\tt B}}
\def\tC{{\tt C}}
\def\tD{{\tt D}}
\def\td{{\tt d}}
\def\tE{{\tt E}}
\def\tte{{\tt e}}
\def\tF{{\tt F}}
\def\tG{{\tt G}}
\def\tg{{\tt g}}
\def\ti{{\tt i}}
\def\tI{{\tt I}}
\def\tj{{\tt j}}
\def\tn{{\tt n}}
\def\tL{{\tt L}}
\def\tO{{\tt O}}
\def\tP{{\tt P}}
\def\tq{{\tt q}}
\def\ttr{{\tt r}}
\def\tP{{\tt P}}
\def\tR{{\tt R}}
\def\tS{{\tt S}}
\def\ttt{\tt t}
\def\tT{{\tt T}}
\def\ttg{{\tt g}}
\def\ttG{{\tt G}}
\def\bttg{\overline{\tg}}
\def\tu{{\tt u}}
\def\tv{{\tt v}}
\def\tV{{\tt V}}
\def\tw{{\tt w}}
\def\tx{{\tt x}}
\def\ty{{\tt y}}
\def\tz{{\tt z}}

\def\bgam{{\mbox{\boldmath$\gamma$}}}
\def\uGam{\underline\Gamma}

\def\boeta{{\mbox{\boldmath$\eta$}}}
\def\oboeta{\overline\boeta}
\def\ups{\upsilon}

\def\Om{\Omega}
\def\om{\omega}
\def\oom{\overline\omega}
\def\bttg{\mbox{\boldmath${\tt g}$}}
\def\btau{\mbox{\boldmath${\tau}$}}
\def\bom{{\mbox{\boldmath$\omega$}}}
\def\obom{\overline\bom}
\def\0bom{{\bom}^0}
\def\0obom{{\obom}^0}
\def\nbom{{\bom}_n}
\def\0nbom{{\bom}_{n,0}}
\def\n*bom{{\bom}^*_{(n)}}
\def\wt{\widetilde}
\def\wtbom{\widetilde\bom}
\def\whbom{\widehat\bom}
\def\oom{\overline\om}
\def\wtom{\widetilde\om}
\def\bOm{\mbox{\boldmath${\Om}$}}
\def\obOm{\overline\bOm}
\def\whbOm{\widehat\bOm}
\def\wtbOm{\widetilde\bOm}

\def\Gam{\Gamma}
\def\Lam{\Lambda}
\def\lam{\lambda}

\def\Ups{\Upsilon}
\def\utheta{\underline\theta}
\def\ovr{\overline r}

\def\oG{\overline G}
\def\oL{\overline L}

\def\bbC{\mathbb C}
\def\bbE{\mathbb E}
\def\bbP{\mathbb P}
\def\fB{\mathfrak B}
\def\fG{\mathfrak G}
\def\fW{\mathfrak W}
\def\bbQ{\mathbb Q}

\def\bi{\mathbf i}
\def\bj{\mathbf j}
\def\bn{\mathbf n}
\def\bt{\mathbf t}
\def\bu{\mathbf u}
\def\bw{\mathbf w}
\def\bX{\mathbf X}
\def\ubX{\underline\bX}
\def\bx{\mathbf x}
\def\ubx{\underline\bx}
\def\bY{\mathbf Y}
\def\by{\mathbf y}
\def\ubY{\underline\bY}
\def\uby{\underline\by}
\def\bZ{\mathbf Z}
\def\bz{\mathbf z}

\def\cl{\centerline}

\def\cA{\mathcal A}
\def\cB{\mathcal B}
\def\cC{\mathcal C}
\def\cD{\mathcal D}
\def\cE{\mathcal E}
\def\cF{\mathcal F}
\def\cH{\mathcal H}
\def\cK{\mathcal K}
\def\cL{\mathcal L}
\def\cN{\mathcal N}
\def\cS{\mathcal S}
\def\ocS{\overline\cS}
\def\cT{\mathcal T}
\def\cV{\mathcal V}
\def\cW{\mathcal W}
\def\ocH{\overline\cH}
\def\ocW{\overline\cW}

\def\bbB{\mathbb B}
\def\bbK{\mathbb K}
\def\bbL{\mathbb L}

\def\bbR{\mathbb R}
\def\bbS{\mathbb S}
\def\bbT{\mathbb T}
\def\bbZ{\mathbb Z}
\def\ba{\mathbf a}
\def\bg{\mathbf g}
\def\bX{\mathbf X}
\def\bx{\mathbf x}
\def\wtbx{\widetilde\bx}
\def\ui{{\underline i}}

\def\oA{{\overline A}}
\def\uA{{\underline A}}
\def\ua{{\underline a}}
\def\uua{{\underline{a_{}}}}
\def\oa{{\overline a}}
\def\uk{{\underline k}}
\def\ux{{\underline x}}
\def\wtux{\widetilde\ux}
\def\uX{{\underline X}}
\def\by{\mathbf y}
\def\uy{\underline y}
\def\bY{\mathbf Y}
\def\uY{\underline Y}

\def\uj{{\underline j}}
\def\unn{\underline n}
\def\unp{\underline p}
\def\ovp{\overline p}
\def\bx{\mathbf x}
\def\ox{\overline x}
\def\obx{\overline\bx}
\def\uz{\underline z}
\def\bz{\mathbf z}
\def\uv{\underline v}
\def\diy{\displaystyle}
\def\ov{\overline}
\def\u0{{\underline 0}}

\def\oomega{\overline\omega}
\def\oUpsilon{\overline\Upsilon}
\def\wtomega{\widetilde\omega}
\def\wtz{\widetilde z}
\def\wtheta{\widetilde\theta}
\def\wtalpha{\widetilde\alpha}
\def\wh{\widehat}
\def\oV{\overline {\mathcal V}}

\def\bI{\mathbf I}
\def\bN{\mathbf N}
\def\bbN{\mathbf N}
\def\bP{\mathbf P}
\def\bV{\mathbf V}
\def\oW{\overline W}
\def\ofW{\overline\fW}
\def\LT{{\mathbb{LT}}}
\def\mucr{{\mu_{cr}}}

\def\rA{{\rm A}}
\def\rB{{\rm B}}
\def\urB{\underline\rB}
\def\rc{{\rm c}}
\def\rC{{\rm C}}
\def\rd{{\rm d}}
\def\rD{{\rm D}}
\def\rd{{\rm d}}
\def\re{{\rm e}}
\def\rE{{\rm E}}
\def\rF{{\rm F}}
\def\rI{{\rm I}}

\def\rn{{\rm n}}

\def\rO{{\rm O}}
\def\rP{{\rm P}}
\def\rQ{{\rm Q}}
\def\rr{{\rm r}}
\def\rR{{\rm R}}

\def\rs{{\rm s}}
\def\rS{{\rm S}}
\def\rT{{\rm T}}
\def\rV{{\rm V}}

\def\rw{{\rm w}}

\def\rx{{\rm x}}
\def\ry{{\rm y}}
\def\rtr{\rm{tr}}

\def\oa{\overline a}
\def\ua{\underline a}

\def\uk{\underline k}
\def\un{\underline n}
\def\ux{\underline x}
\def\uy{\underline y}
\def\wtux{\widetilde\ux}
\def\uX{\underline X}

\def\oJ{\overline J}
\def\oP{\overline P}
\def\utC{{\underline\tC}}
\def\utD{{\underline\tD}}
\def\utE{{\underline\tE}}
\def\urB{{\underline\rB}}
\def\urC{{\underline\rC}}
\def\urD{{\underline\rD}}
\def\urE{{\underline\rE}}
\def\vng{{\varnothing}}
\def\ueta{\underline{\eta}}
\def\wt{\widetilde}
\def\fB{\mathfrak B}\def\fM{\mathfrak M}\def\fX{\mathfrak X}
 \def\cB{\mathcal B}\def\cM{\mathcal M}\def\cX{\mathcal X}
\def\mbe{\mathbf e}
\def\bu{\mathbf u}\def\bv{\mathbf v}\def\bx{\mathbf x} \def\by{\mathbf y} \def\bz{\mathbf z}
\def\om{\omega} \def\Om{\Omega}
\def\bbP{\mathbb P} \def\hw{h^{\rm w}} \def\hwi{{h^{\rm w}}}
\def\beq{\begin{eqnarray}} \def\eeq{\end{eqnarray}}
\def\beqq{\begin{eqnarray*}} \def\eeqq{\end{eqnarray*}}
\def\rd{{\rm d}} \def\Dwphi{{D^{\rm w}_\phi}}
\def\BX{\mathbf{X}}\def\Lam{\Lambda}\def\BY{\mathbf{Y}}
\def\BZ{\mathbf{Z}} \def\BN{\mathbf{N}}
\def\BV{\mathbf{V}}

\def\mwe{{D^{\rm w}_\phi}}
\def\DwPhi{{D^{\rm w}_\Phi}} \def\iw{i^{\rm w}_{\phi}}
\def\bE{\mathbb{E}}
\def\1{{\mathbf 1}} \def\fB{{\mathfrak B}}  \def\fM{{\mathfrak M}}
\def\diy{\displaystyle} \def\bbE{{\mathbb E}} \def\bu{\mathbf u}
\def\BC{{\mathbf C}} \def\lam{\lambda} \def\bbB{{\mathbb B}}
\def\bbR{{\mathbb R}}\def\bbS{{\mathbb S}}
 \def\bmu{{\mbox{\boldmath${\mu}$}}}
 \def\bPhi{{\mbox{\boldmath${\Phi}$}}}  \def\bPi{{\mbox{\boldmath{$\Pi$}}}}
 \def\bbZ{{\mathbb Z}} \def\fF{\mathfrak F}\def\mbt{\mathbf t}\def\B1{\mathbf 1}
\def\hwphi{h^{\rm w}_{\phi}}
\def\BW{\mathbf{W}} \def\bw{\mathbf{w}}

\def\beal{\begin{array}{l}}
\def\beac{\begin{array}{c}}
\def\beacl{\begin{array}{cl}}
\def\ena{\end{array}}
\def\WBJ{\mathbf{J}^{\rm w}_{\phi}}
\def\BS{\mathbf{S}}
\def\BK{\mathbf{K}}
\def\BB{\mathbf{B}}
\def\wtD{{\widetilde D}}

\def\mwe{{D^{\rm w}_\phi}}
\def\DwPhi{{D^{\rm w}_\Phi}} \def\iw{i^{\rm w}_{\phi}}
\def\bE{\mathbb{E}}
\def\1{{\mathbf 1}} \def\fB{{\mathfrak B}}  \def\fM{{\mathfrak M}}
\def\diy{\displaystyle} \def\bbE{{\mathbb E}} \def\bu{\mathbf u}
\def\BC{{\mathbf C}} \def\lam{\lambda}
\def\bbB{{\mathbb B}} \def\bbM{{\mathbb M}}
\def\bbR{{\mathbb R}}\def\bbS{{\mathbb S}}
\def\blam{{\mbox{\boldmath${\lambda}$}}}
\def\bmu{{\mbox{\boldmath${\mu}$}}} \def\bta{{\mbox{\boldmath${\eta}$}}}
\def\bzeta{{\mbox{\boldmath${\zeta}$}}}
 \def\bPhi{{\mbox{\boldmath${\Phi}$}}}  \def\bPi{{\mbox{\boldmath{$\Pi$}}}}
 \def\bbZ{{\mathbb Z}} \def\fF{\mathfrak F}\def\mbt{\mathbf t}\def\B1{\mathbf 1}
\def\hwphi{h^{\rm w}_{\phi}}
\def\BT{{\mathbf T}} \def\BW{\mathbf{W}} \def\bw{\mathbf{w}}
\def\bfe{{\mathbf e}}
\def\beps{{\mathbf \varepsilon}}

\def\beal{\begin{array}{l}}
\def\beac{\begin{array}{c}}
\def\beacl{\begin{array}{cl}}
\def\ena{\end{array}}
\def\WBJ{\mathbf{J}^{\rm w}_{\phi}}
\def\BS{\mathbf{S}}
\def\BK{\mathbf{K}}
\def\tL{\mathbf{L}}
\def\BB{\mathbf{B}}
\def\vphi{{\varphi}}
\def\rw{{\rm w}}
\def\bZ{\mathbf Z}
\def\wtf{{\widetilde f}} \def\wtg{{\widetilde g}} \def\wtG{{\widetilde G}}
\def\vphi{\varphi}
\def\rT{{\rm T}}
\def\tA{{\tt A}} \def\tB{{\tt B}} \def\tC{{\tt C}} \def\tI{{\tt I}} \def\tJ{{\tt J}} \def\tK{{\tt K}}
\def\tL{{\tt L}} \def\tP{{\tt P}} \def\tQ{{\tt Q}} \def\tS{{\tt S}}
\def\beac{\begin{array}{c}} \def\beal{\begin{array}{l}} \def\beacl{\begin{array}{cl}} \def\ena{\end{array}}
\def\bbV{\mathbb{V}}
\def\bbS{\mathbb{S}}

\def\bu{\mathbf{u}} \def\BU{\mathbf{U}}
\def\bbN{\mathbb{N}}

%\hyphenation{op-tical net-works semi-conduc-tor}

% TODO Where's the right place to put this now?
\begin{figure}[h]
    \centering
    \includegraphics[width=\columnwidth]{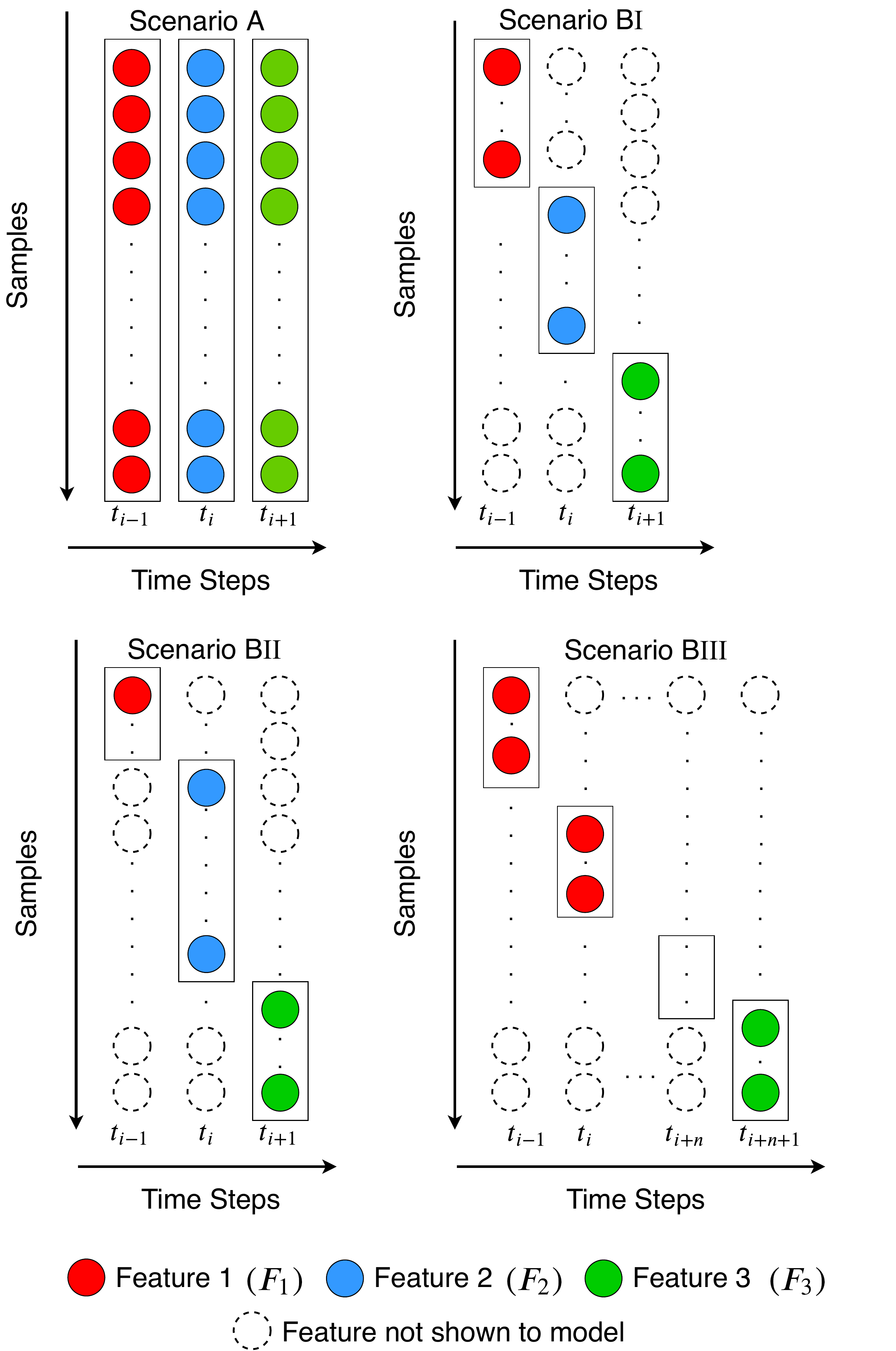}
    \caption{Illustration of the scenarios applicable to GOA. Scenario A represents the conventional OSFS setting where a data from all samples are used in a time step. Scenarios B represent the novel setting introduced in this paper, when both samples and features are streaming. (B\RNum{1}) The number of samples across each time step are equal in size and each time step introduces a new feature. (B\RNum{2}) The number of samples across each time step are unequal in size and each time step introduces a new feature. (B\RNum{3}) The number of samples across each time step are equal in size and each feature can span multiple time steps.}
    \label{fig.Scenarios}
\end{figure}

\section{Approach}
\label{sec:algs}
\subsection{Setup: OSFS with Streaming Samples}
 \label{subsec:osfsss}
   Given class label $C$, a new incoming feature $F_i$, a threshold $\delta\in[0,1]$, the selected feature set at time $t_{i-1}$, $S^*_{t_{i-1}}$, and the current time step $t_{i}$, the observed streaming sample matrix with feature set $S^*_{t_i-1}\cup F_i$ at time $t_i$ is $\mathcal{A}_{t_i}$.
 
Under this proposed OSFS-SS setup we consider four distinct scenarios (Fig.~\ref{fig.Scenarios}):
 \begin{itemize}
     \item {\bf Scenario A :} Traditional OSFS setting, where at each time step $t_i$, only one new incoming feature $F_i$ for all samples in the dataset is observed.
     \vspace{0.2cm}
     \item {\bf Scenario B\RNum{1} :} At time step $t_i$, only one new incoming feature $F_i$ for a subset of samples $\{x_1,x_2,\ldots,x_M\}$ is observed. 
     The number of samples across different time steps are equal.
       \vspace{0.2cm}
     \item {\bf Scenario B\RNum{2} :} At time step $t_i$, only one new coming feature $F_i$ along with the associated subset of samples\\
     $\{x_1,x_2,\ldots,x_{M_i}\}$ is observed. However, the number of samples across different time steps are not equal.
       \vspace{0.2cm}
     \item {\bf Scenario B\RNum{3} :} This scenario is an extended version of Scenario B\RNum{1} where we may observe the same feature $F_i$ for more than one time step. However, across time steps different sets of samples are observed. 
 \end{itemize}
\subsection{Geometric Online Adaptation}
 \label{subsec:goa}
  \subsubsection{Review of CGD}
 \label{subsec:cgd}
 \hfill\\
 
 Let us recall the conditional geometric dependency measure proposed in \cite{ICASSPsalimeh2019}, denoted by $G_\rho$, to evaluate dependencies between features given a class label $C$.

 {\bf Definition:} (Conditional Geometric Dependency) Let $X$ and $Y$ be two given features with joint probability density function $f(x,y)$. Consider class label variable $C$ taking values in $\{1,2\ldots,m\}$. Let $p_i=P(C=i)$ for $i=1,\ldots,m$ such that $\diy\sum\limits_{i=1}^m p_i=1$. 
 Then the conditional geometric dependency measure $G_{\rho}:=G_\rho(X,Y|C)$ is given by
 \begin{equation}
 G_{\rho}=1-2\sum\limits_i\sum\limits_j p_i p_j\eta_{ij}.
 \end{equation}
 where $\eta_{ij}$ is the following measure based on joint probability densities $f(x,y)$ and $\pi(x,y)$:
 \begin{equation}\label{etaij}
   \eta_{ij}=\diy\iint \frac{f_{X,Y|C}(x,y|i)f_{X|C}(x|j)f_{Y|C}(y|j)}{f(x,y)+\pi(x,y)} \;{\rm d} x {\rm d} y. 
 \end{equation}
 and 
 \begin{equation}\begin{array}{cl}
 f(x,y)= \diy\sum\limits_{i=1}^m p_i f_{X,Y|C}(x,y|i),\\ \nonumber
 \pi(x,y)=\diy\sum\limits_{i=1}^m p_i f_{X|C}(x|i) f_{Y|C}(y|i),
 \end{array}\end{equation}
 such that $f_{X,Y|C}(x,y|i)$, $f_{X|C}(x|i)$, and $f_{Y|C}(y|i)$ are the conditional distribution of conditional random variable $X,Y|C=i$, $X|C=i$, and $Y|C=i$ respectively. 

 {\bf $G_\rho$ Estimator:} Following arguments in \cite{SalimehEntropy2018} and \cite{ICASSPsalimeh2019}, in this paper we employ a graph-based estimation of $G_\rho$ denoted by $\widehat{G}_\rho$. This estimator is computed by the global Friedman-Rafsky (FR) multivariate run test statistic constructed using a global minimal spanning tree (MST). This approach results in an efficient and fast non-parametric implementation of the conditional geometry dependency estimation.

\subsubsection{Algorithm Description}\hfill\\
 Our proposed Geometric Online Adaptation (GOA) approach (Alg.~\ref{algorithm0}) functions in the context of Scenario B\RNum{1} as described below. The algorithm can be extended to the two other streaming Scenarios B\RNum{2} and B\RNum{3} by selecting the streaming sample block that has the highest $\widehat{G}_\rho(F_i;X|C)$ for an incoming feature $F_i$, where $X\in S^*_{t_{i-1}}$. 

 \begin{itemize}
 \setlength{\itemsep}{0pt}
 \setlength{\parskip}{0pt}
 \setlength{\parsep}{0pt}
     \item We first compute $\widehat{G}_\rho(F_i;X|C)$, $\widehat{G}_\rho(F_i;Z|C)$, and $\widehat{G}_\rho(X;Z|C)$ for new coming feature $F_i$ and features $X,Z\in S^*_{t_{i-1}}$ using $\mathcal{A}_{t_i}$.
     \item $F_i$ is discarded if $\widehat{G}_\rho(F_i;X|C)\geq \delta$ and $\widehat{G}_\rho(F_i;Z|C)>\linebreak \widehat{G}_\rho(X;Z|C)$, otherwise it is merged in to set $S^*_{t_{i-1}}$ as $S^*_{t_i}=S^*_{t_{i-1}}\cup F_i$. Note that $S_{t_i}$ is still not the selected subset of features at time $t_i$ and needs to be evaluated further. 
     \item If $F_i$  is merged in $S^*_{t_{i-1}}$, the selected feature set $S^*_{t_{i}}$ is computed. For this, the set $S^*_{t_{i-1}}$ is updated based on the relationship between $F_i$ and features in $S^*_{t_{i-1}}$. Given class variable $C$ for $Z,X, Y\in S^*_{t_{i}}$, $X\neq Y\neq Z$ if $\widehat{G}_{\rho}(Z;X|C) > \delta$, and $\widehat{G}_{\rho}(Z;X|C)\geq \widehat{G}_{\rho}(Y;X|C)$ then $Z$ is removed from $S^*_{t_{i-1}}$. Thus $S^*_{t_{i}}=S_{t_{i}}-Z$. This is because feature $Z$ does not increase the predictive capability. 
     \item These steps are repeated for each new incoming feature with new samples.
 \end{itemize}
 In the proposed GOA method the criterion $G_{\rho}(F_i;X|C)\geq \delta$ can be motivated as follows:  Given conditional geometric mutual information (MI) $I(F_i;X|C)$ proposed in \cite{SalimehEntropy2018}, \cite{ICASSPsalimeh2019}:
 \begin{equation}\label{def.CMI}\begin{array}{l}
     I(F_i;X|C)=\\[10pt]
     1- 2\diy\sum\limits_{c}f(c)\iint \diy\frac{f(F_i,x|c) f(F_i|c)f(x|c)}{f(F_i,x|c)+f(F_i|c)f(x|c)}\; \rd F_i\; \rd x, 
\end{array} \end{equation}
 and the fact that $I(F_i;X|C)\geq G_{\rho}(F_i;X|C)$ \cite{ICASSPsalimeh2019}, the criterion $\linebreak G_{\rho}(F_i;X|C)\geq \delta$ implies $I(F_i;X|C)\geq \delta$. Now if $\exists X\in S^*_{t_{i-1}}$ such that $I(F_i;C|X)=0$ then adding $F_i$ to $S^*_{t_{i-1}}$ does not improve subset $S^*_{t_{i-1}}$ in terms of prediction accuracy because $F_i$ and class label variable $C$ are independent given $X$. This result is stated in the next section.

 \begin{algorithm}[h]\label{algorithm1}
  \caption{The GOA Algorithm with Streaming Samples}
  \begin{algorithmic}
  \renewcommand{\algorithmicrequire}{\textbf{Input:}}
  \renewcommand{\algorithmicensure}{\textbf{Output:}}
  \REQUIRE $F_i$: predictive feature; $C$: the class labels, $0\leq\delta\leq 1$: a relevance threshold; \\$S^*_{t_{i-1}}$: the selected feature set at time $t_{i-1}$;
  $S^*_{t_{i}}$: the selected feature set at time $t_{i}$;
  $n_{t_i}$: the number of sample at time $t_i$;
 % $\{\bx_1,\bx_2,\ldots,\bx_{n_{t_i}}\}$: the observed streaming sample at time $t_i$;
  \\$\mathcal{A}_{t_i}$: the observed streaming sample matrix with feature set $S^*_{t_i-1}\cup F_i$ at time $t_i$;
%   \vspace{0.2cm}
   \STATE {\bf Repeat}
   \STATE Get a new sample  $\mathcal{A}_{t_i}$ at time $t_i$, which includes a new  \\feature $F_i$;
   \STATE \quad{\bf for} features $Z,X\in S^*_{t_{i-1}}$, $Z\neq X$ {\bf compute}
   \STATE \quad \quad $\widehat{G}_{\rho}(F_i;X|C)$, $\widehat{G}_{\rho}(F_i;Z|C)$, $\widehat{G}_{\rho}(X;Z|C)$ using $\mathcal{A}_{t_i}$\\
   \STATE \quad \quad and {\bf do}
     \STATE \quad{\bf if} $\widehat{G}_{\rho}(F_i;X|C)\geq \delta$  and $\widehat{G}_{\rho}(F_i;Z|C) > \widehat{G}_{\rho}(X;Z|C)$\\
     \qquad{\bf then} Discard $F_i$ and go to Step 11
   \STATE \quad{\bf else}  $S^*_{t_{i}}=S^*_{t_{i-1}}\cup F_i$
   \STATE \quad{\bf end}
   \STATE \quad{\bf for} features $Z,Y,X\in S^*_{t_{i}}$, $Z\neq Y\neq X$ {\bf do}
     \STATE \quad{\bf if} $\widehat{G}_{\rho}(Z;X|C) > \delta$, and $\widehat{G}_{\rho}(Z;X|C)\geq \widehat{G}_{\rho}(Y;X|C)$\\
     \qquad{\bf then} $S^*_{t_{i}}=S^*_{t_{i}}-Z$
     \STATE {\bf until} no features and sample are streaming;
   \ENSURE  $S^*_{t_{i}}$: the selected feature set
  \end{algorithmic} 
 \label{algorithm0}
\end{algorithm}

When comparing the formulation and assumptions of GOA (Alg.~\ref{algorithm0}) to its closest competitor, SAOLA, we observe that SAOLA discards an incoming feature by comparing its relevancy to the class labels while in our method we incorporate dependency between features given the class label.
Using this approach, we reduce the effective number of steps from two in SAOLA to a single step in GOA, as well as the effective number of computations. using the geometric dependency measure.
The most important reason we were able to achieve this is by employing the geometry of samples.
Also, by using the geometric dependency criterion, the MI estimate is truly bounded between 0 and 1 while it is not the case in SAOLA.

  %-----------------------------------------
  \def\bbE{\mathbb{E}}

 \subsubsection{Theoretical Analysis}\hfill\\
 Given class variable $C$ and features $X$ and $F_i$, the theorem below shows the existence of a lower bound for $I(F_i;X|C)$ when $I(F_i;C|X)=0$. The proof is provided here followed by two required Lemmas \ref{lemma.1} and \ref{lemma.2}.
 \begin{theorem}
 \label{thm.1}
 Given the current feature subset $S^*_{t_{i-1}}$ at time $t_{i-1}$ and a new feature $F_i$ at time $t_i$, if $\exists X\in S^*_{t_{i-1}}$, such that $I(F_i;C|X)=0$ then $I(F_i;X|C)\geq {\delta}$, where ${\delta}\in(0,1)$.
 \end{theorem}
 {\bf Proof of Theorem 3.1:} The following two lemmas are required to prove Theorem \ref{thm.1}:

 \begin{lemma}\label{lemma.1}
 Let $S^*_{t_{i-1}}$ be current feature subset at time $t_{i-1}$ and $F_i$ denotes the new feature at time $t_i$. For $X\in S^*_{t_{i-1}}$, we have
 \begin{equation}\label{eq.0}
 I(F_i;X|C)\geq I(F_i,C;X)-2\bbE_f\left[\left(\frac{f(X)}{f(X|C)}+1\right)^{-1}\right].
 \end{equation}
 \end{lemma}
 {\bf Proof of Lemma 3.2:}
Given $p\in(0,1)$ and $q=1-p$, we can easily check that for positive $t>q$, $s>q$, such that $t,s\neq 1$:
$$(t+s)\left[(ts)+q(1-t-s)\right]-p (ts)\geq 0.$$
This implies
\begin{equation}\label{eq.1}
p\Big(\diy\frac{t-q}{p}\Big)\Big(\frac{s-q}{p}\Big)+q\geq \diy\frac{t\;s}{t+s}.\end{equation}
In (\ref{eq.1}), substitute 
$$\frac{s-q}{p}=\frac{f(F_i,c;x)}{f(F_i,c) f(x)},\;\; \frac{t-q}{p}=\frac{f(x)}{f(x|c)},$$
therefore the LHS in (\ref{eq.1}) becomes 
$$p\left(\frac{f(F_i,c;x)}{f(F_i,c) f(x)}\right)\left(\frac{f(x)}{f(x|c)}\right)+q=p\left(\frac{f(F_i,x|c)}{f(F_i|c)f(x|c)}\right)+q.$$
Set $p=q=1/2$, hence by conditional mutual information definition in (\ref{def.CMI}), we have 
\begin{equation}\label{eq.2}
I(F_i;X|C)\geq 1-\bbE_f\left[\frac{1}{s}\right]-\bbE_f\left[\frac{1}{t}\right],
\end{equation}
where $1-\diy\bbE_f\left[s^{-1}\right]=I(F_i,C;X)$ and
\begin{equation}\label{eq.3}
\bbE_f\left[t^{-1}\right]=2\bbE_f\left[\left(\frac{f(X)}{f(X|C)}+1\right)^{-1}\right].
\end{equation}
Combine (\ref{eq.2}) and (\ref{eq.3}). This implies our claim inequality (3) in Lemma 3.2. 
 %-------------------------------------------
 \begin{lemma}\label{lemma.2}
 As in Lemma \ref{lemma.1} with the current feature subset $S^*_{t_{i-1}}$ at time $t_{i-1}$ and a new feature $F_i$ at time $t_i$, if $\exists X\in \bbS^*_{t_{i-1}}$ such that 
 \begin{equation}\label{eq.5}
 f(F_i|x)f(c|x)\geq f(F_i)f(c),    
 \end{equation}
 then $\exists \tilde{\delta}\geq 0$ such that 
 \begin{equation}\label{lemma2.eq}
 I(F_i,C;X)\geq I(F_i;C|X)+ \tilde{\delta}.
 \end{equation}
 \end{lemma}
 
 {\bf Proof of Lemma 3.3:} 
From assumption (7), we can get that $\exists \bar{\delta}\geq 0$ such that for all $\epsilon\geq 0$
\begin{equation}
\left(\frac{1}{f(F_i|x)f(c|x)}+\epsilon\right)^{-1}-\left(\frac{1}{f(F_i)f(c)}+\epsilon\right)^{-1}\geq \bar{\delta}.
\end{equation}
Now set $\epsilon=\left(f(F_i,c|x)\right)^{-1}$, then $\exists \tilde{\delta}\geq 0$ such that
\begin{equation}\label{eq.6}
\bbE_f\left[\left(\frac{f(F_i,C|X)}{f(F_i|X)f(C|X)}+1\right)^{-1}-\left(\frac{f(F_i,C|X)}{f(F_i)f(C)}+1\right)^{-1}\right]\geq \tilde{\delta}.
\end{equation}
On the other hand we discard $F_i$ if $f(c|F_i)=f(c)$ and $f(c|x,F_i)=f(c|x)$ for $x\in S^*_{t_{i-1}}$. This implies $f(F_i,c)=f(F_i)f(c)$, therefore the LHS in (\ref{eq.6}) becomes
\begin{equation}\label{eq.7}
\bbE_f\left[\left(\frac{f(F_i,C|X)}{f(F_i|X)f(C|X)}+1\right)^{-1}-\left(\frac{f(F_i,C,X)}{f(F_i,C)f(X)}+1\right)^{-1}\right]
\end{equation}
The term (\ref{eq.7}) equals $I(F_i,C;X)-I(F_i;C|X)$, consequently going back to (\ref{eq.6}), $\exists {\tilde{\delta}}\geq 0$ such that
\begin{equation}
I(F_i,C;X)-I(F_i;C|X)\geq \tilde{\delta}.
\end{equation}
This completes the proof of Lemma 3.3. 
 
 %------------------------------------------
 Now, by combining (\ref{eq.0}) and (\ref{lemma2.eq}) we derive our claim in Theorem \ref{thm.1} i.e. $I(F_i;X|C)\geq \delta$. \hfill $\square$\\
 
From Theorem \ref{thm.1}, we infer that if there exists $X\in S^*_{t_{i-1}}$ such that $I(F_i;X|C)\leq \delta$ then there is sufficient useful information stored in feature $F_i$ that is important for predicting class variable $C$.
 Moreover, there isn't too much overlap between $F_i$ and $X$.
 We evaluate this by searching all features $X\in S^*_{t_{i-1}}$ and if $I(F_i;X|C)\geq \delta$ then given $X$, $F_i$ is independent from $C$ and we can discard $F_i$. This is our approach in Algorithm~\ref{algorithm0}.

From the above discussion, one can intuitively infer the following conjecture.\\ 
 {\bf Conjecture:} Given the current feature subset $S^*_{t_{i-1}}$ at time $t_{i-1}$ and a new feature $F_i$ at time $t_i$, if $\exists X,Z\in S^*_{t_{i-1}}$, $X\neq Z$ such that $I(F_i;C|Z)=0$ and $I(F_i;C|X,Z)=0$ then we have $I(F_i;Z|C)\geq I(X;Z|C)$. Further discussion on this is provided in supplementary material. 

  %-----------------------------------------------
 
\subsection{Fair Comparison against Baseline} 
 To ensure fair performance comparisons across baselines and datasets we choose to constrain our algorithm to always use $\delta$ values that correspond to smaller feature subsets than the baseline against which GOA is being compared, i.e., the GOA results are to be comparatively assessed w.r.t. a baseline. 
 More specifically, we perform a grid search of $\delta$ and select the case which corresponds to the highest accuracy when the selected number of features is lower than or equal to the number of features selected by the baseline being compared against.
 In case GOA does not select a lesser number of features than the baseline, we select the hyper-parameter which corresponds to the closest number of features greater than the baseline, regardless of the performance.

\section{Experiments}
\label{sec:experiments}

 Feature selection, given streaming samples and features, is a relatively new problem domain with minimal previous work.
  In an effort to provide more comparable baselines and explore the strengths of our proposed GOA algorithm we divide experiments into three subsections: 1) Traditional OSFS where the total number of samples remain fixed and features are streaming, 2) OSFS-SS on common machine learning datasets where both samples and features are streaming simultaneously, and 3) OSFS-SS applied to a variety of CNN-based network features.
  Since OSFS is a standard problem domain, all the baselines listed below are applicable to this setting while only the extended SAOLA algorithm (X-S), our proposition, is applicable to the OSFS-SS setting.
  All reported accuracies are computed across 5 trials which randomly permute the data matrix, when a fixed train/test set doesn't exist.

  {\bf Baselines:} We compare the proposed GOA algorithm to three state-of-the-art baselines: Alpha-investing \cite{Zhouetal2005}, Fast-OSFS \cite{Wuetal.2013}, and SAOLA \cite{Yuetal2014}. 
  Each of these methods provide their selected feature subsets which are then used by two multi-class classification methods, KNN, and linear SVM, to predict the testing accuracy.

%  {\bf Tuning Threshold $\delta$:} 
%  To ensure fair performance comparisons across baselines and datasets we choose to constrain our algorithm to always use $\delta$ values that correspond to smaller feature subsets than the baseline against which GOA is being compared, i.e., the GOA results are to be comparatively assessed w.r.t. a baseline. 
%  More specifically, we perform a grid search of $\delta$ and select the case which corresponds to the highest accuracy when the selected number of features is lower than or equal to the number of features selected by the baseline being compared against.
%  In case GOA does not select a lesser number of features than the baseline, we select the hyper-parameter which corresponds to the closest number of features greater than the baseline, regardless of the performance.

 {\bf Simulation Datasets:} For our first experiment on OSFS, we setup simulation case studies as multi-class classification problems using samples from multivariate gaussian distributions. 
 Class means are randomly selected from a hyper-sphere of radius 2 units. 
 Multivariate Gaussian distributions with the same randomly generated covariance matrix, $\Sigma$, are centered over each class mean. 
 We perform three studies to measure the impact of, 1) sample size per class, 2) dimensionality, and 3) number of classes on the performance of streaming feature selection baselines.

 {\bf Real-world Datasets:}
 To illustrate the potency of our proposed approach we perform a pairwise comparison against baselines over real-world datasets. 
 Apart from the standard datasets used in literature, we use CNN-based features for the image dataset CIFAR-10.
 We note that the deep learning features for CIFAR-10 were extracted from the second last layer across AlexNet~\cite{krizhevsky2012imagenet}, VGG16~\cite{simonyan2014very}, ResNet18 and ResNet34~\cite{he2016deep}, all of which were initialized with ImageNet pre-trained weights.
 Table~\ref{table.data} lists all the real-world datasets, GESTURE~\cite{madeo2013gesture}, WDBC~\cite{mangasarian1990cancer}, Fashion MNIST~\cite{xiao2017/online}, MNIST~\cite{lecun1998gradient}, CIFAR-10~\cite{krizhevsky2009learning}, Madelon~\cite{NIPS2004_2728}, and MFEAT~\cite{van1998handwritten} and their key characteristics.
 We note that CNN-based features were further reduced using PCA to 310 principal components so as to enable streaming across features and samples.

 % \usepackage{booktabs}
 \begin{table}[]
 \caption{List of real-world datasets used in our experiments and a summary of their characteristics. F-MNIST is Fashion MNIST.}
 \label{table.data}
 \centering
 \begin{tabular}{@{}lrrrr@{}}
 \toprule
 Dataset & \# Features & \# Train & \# Test & \#Labels \\ \midrule
 GESTURE & 19          & 3000     & 5828    & 5        \\
 WDBC    & 32          & 398      & 171     & 2        \\
 F-MNIST & 50          & 60000    & 10000   & 10       \\
 MNIST   & 50          & 60000    & 10000   & 10       \\
 CIFAR10 & 310         & 50000    & 10000   & 10       \\ 
 Madelon & 500         & 2000     & 600     & 2        \\
 MFEAT   & 649         & 1400     & 600     & 10       \\\bottomrule
 \end{tabular}
 \end{table}

%  \begin{figure*}
%       \centering
%       \subfloat[]{\includegraphics[width=0.69\columnwidth]{Figs/FS-AlphavsGOA.pdf}}    
%       \subfloat[]{\includegraphics[width=0.69\columnwidth]{Figs/FS-FastvsGOA.pdf}}    
%       \subfloat[]{\includegraphics[width=0.69\columnwidth]{Figs/FS-SAOLAvsGOA.pdf}}    
%       \caption{Comparison of the number of selected features between the proposed GOA algorithm with ALPHA-investing ("ALPHA"), Fast-OSFS ("FAST"), and SAOLA using KNN and SVM classifiers.}
%       \label{figfeatures}
%   \end{figure*}

 \begin{figure*}
      \centering
      \subfloat[\label{fig:samplesize}]{\includegraphics[width=0.67\columnwidth]{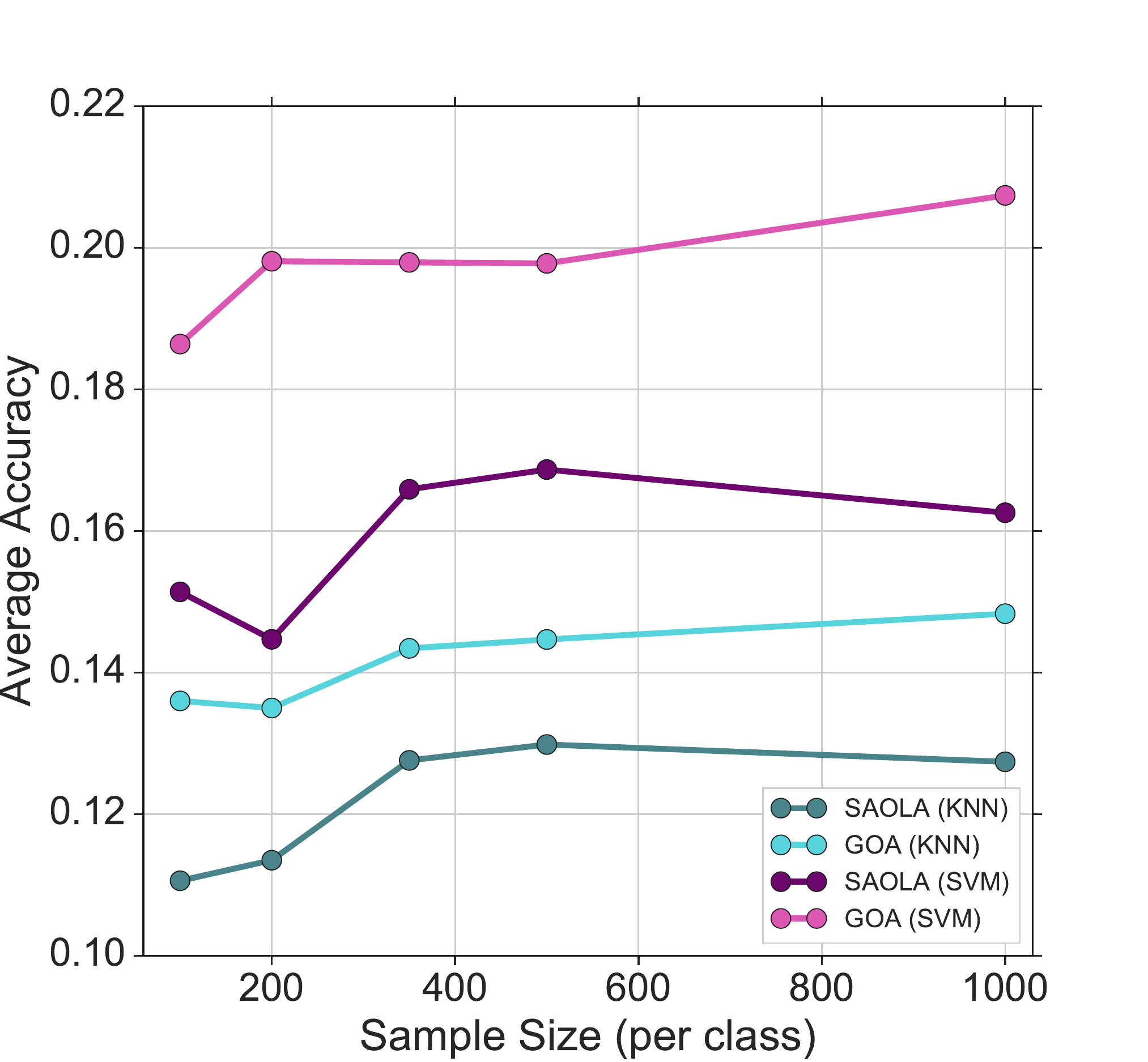}} 
      \subfloat[\label{fig:dimsize}]{\includegraphics[width=0.67\columnwidth]{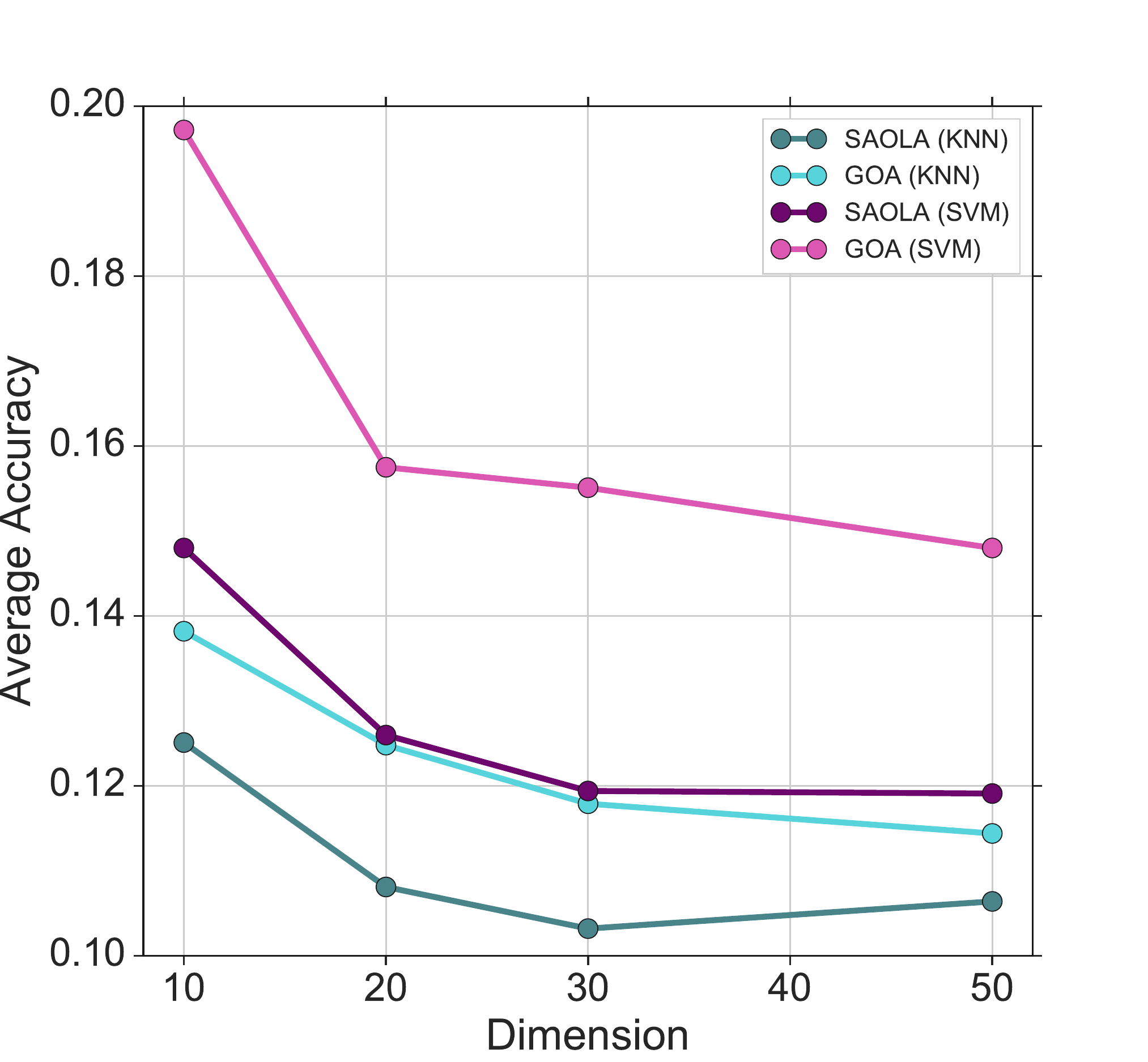}}    
      \subfloat[\label{fig:classsize}]{\includegraphics[width=0.67\columnwidth]{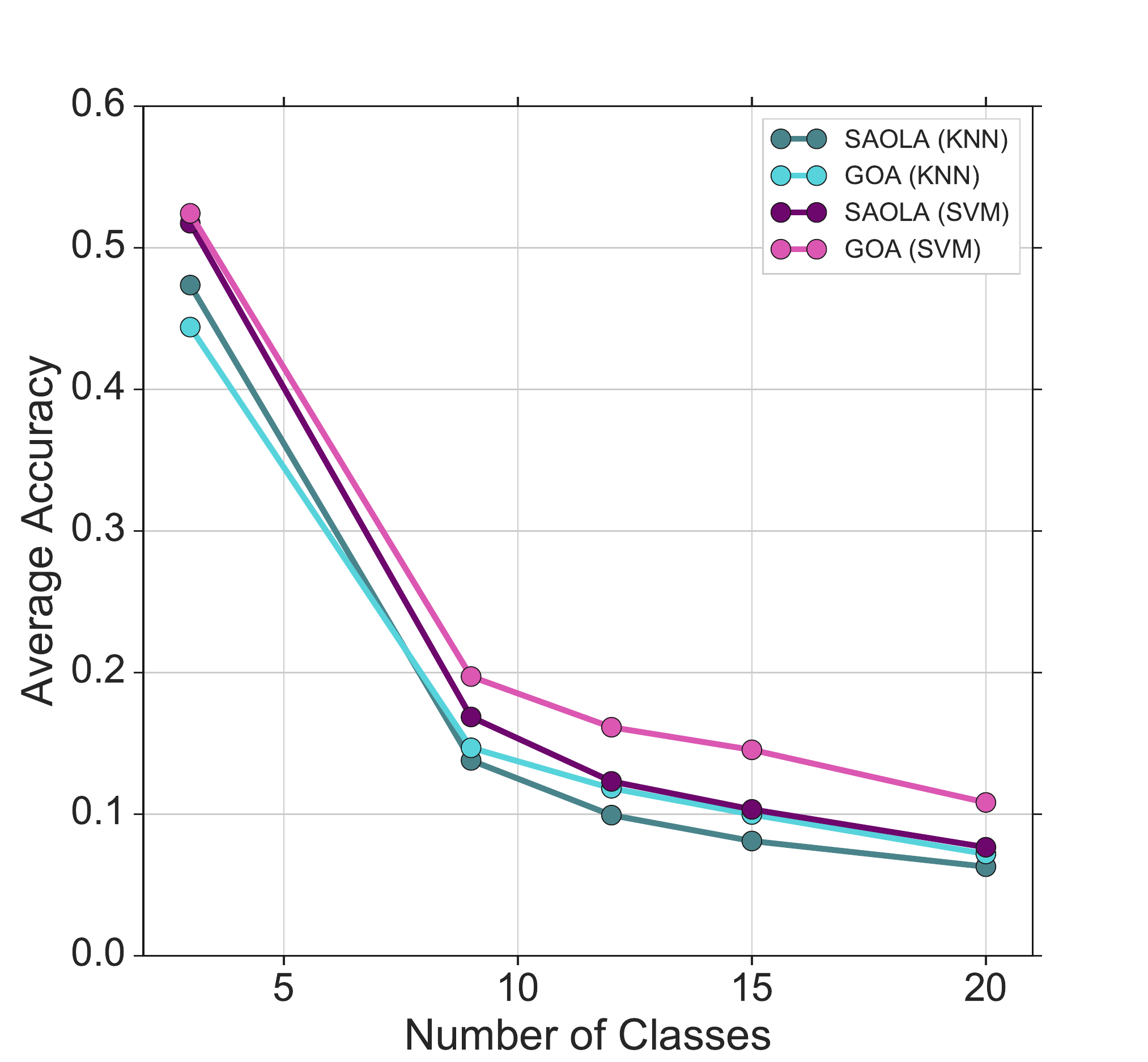}}    
      \caption{Illustration of results from studies measuring the impact of (a) samples per class $[10^2, 10^3]$, (b) dimensionality of features $[10,50]$, and (c) number of classes $[3-20]$. Average prediction accuracy for GOA and SAOLA using both KNN and linear SVM classifiers are shown.
      GOA clearly outperforms SAOLA across each case study. More importantly, the influence of $\widehat{G}_\rho$ is apparent in the results of GOA, illustrated by a slight dip in performance when dimensionality of the feature vector or the number of classes is increased. 
      }
      \label{sim_results}
  \end{figure*}
  
  \begin{figure*}
      \centering
      \subfloat[]{\includegraphics[width=0.33\textwidth]{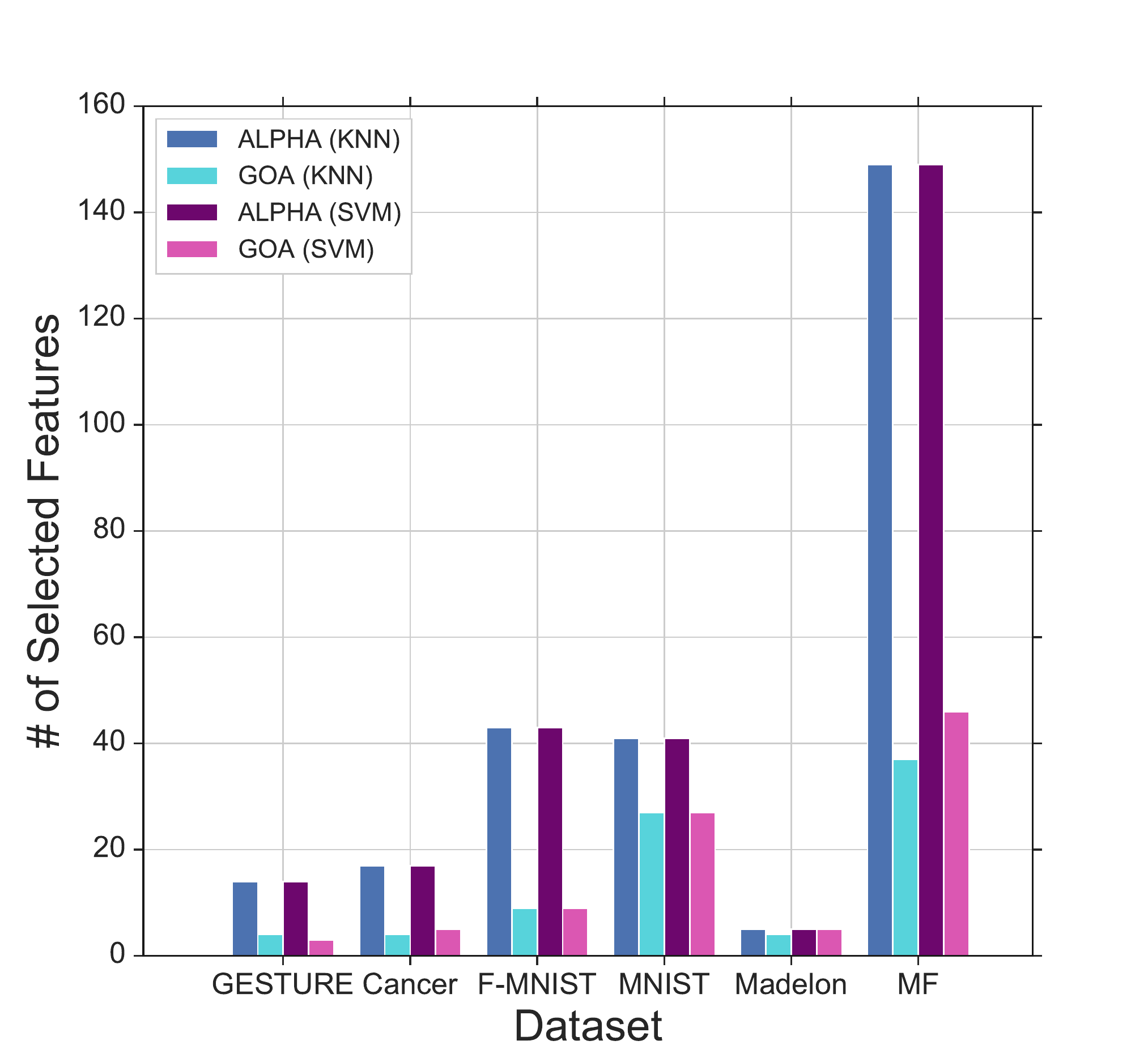}}    
      \subfloat[]{\includegraphics[width=0.33\textwidth]{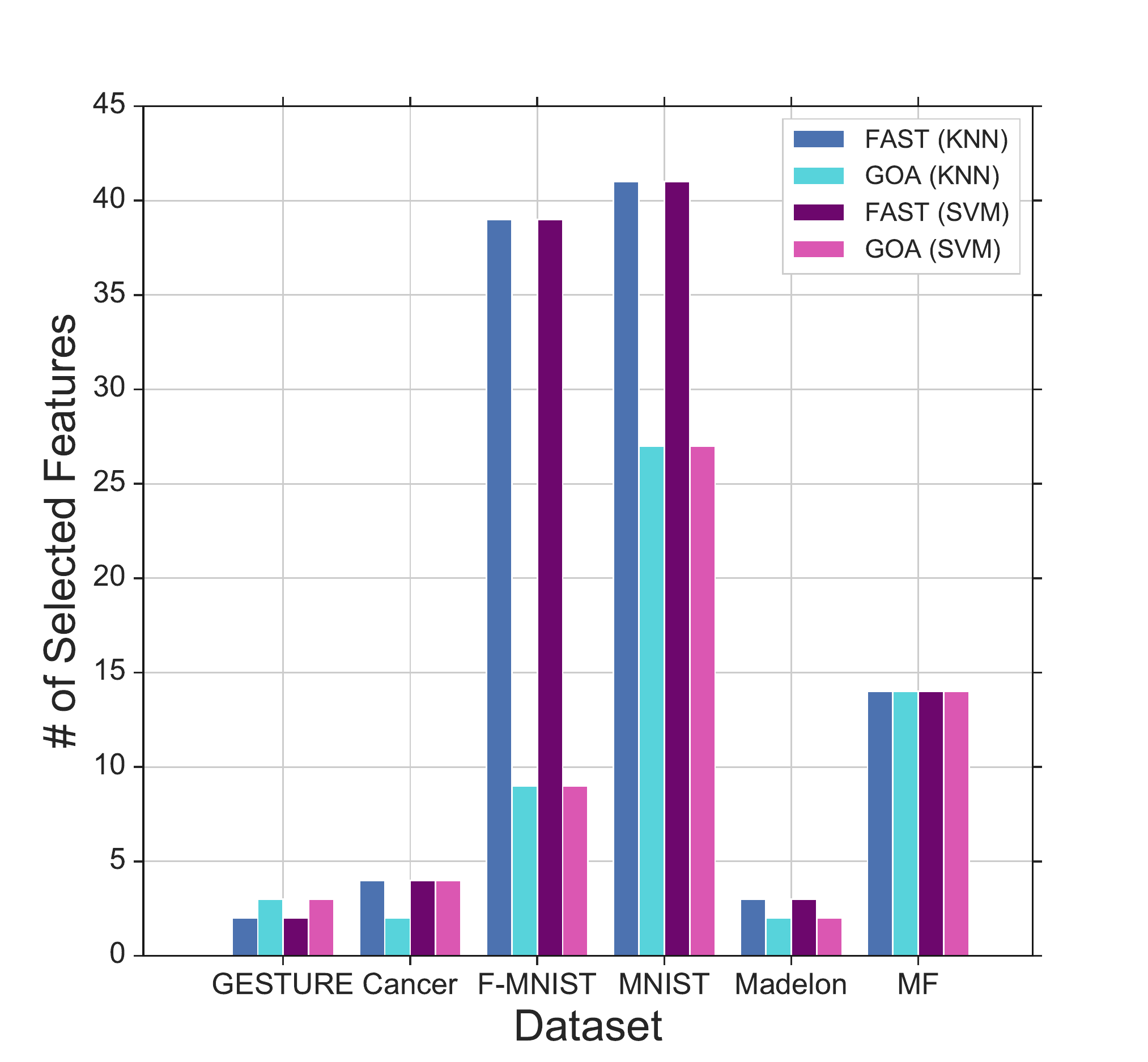}}    
      \subfloat[]{\includegraphics[width=0.33\textwidth]{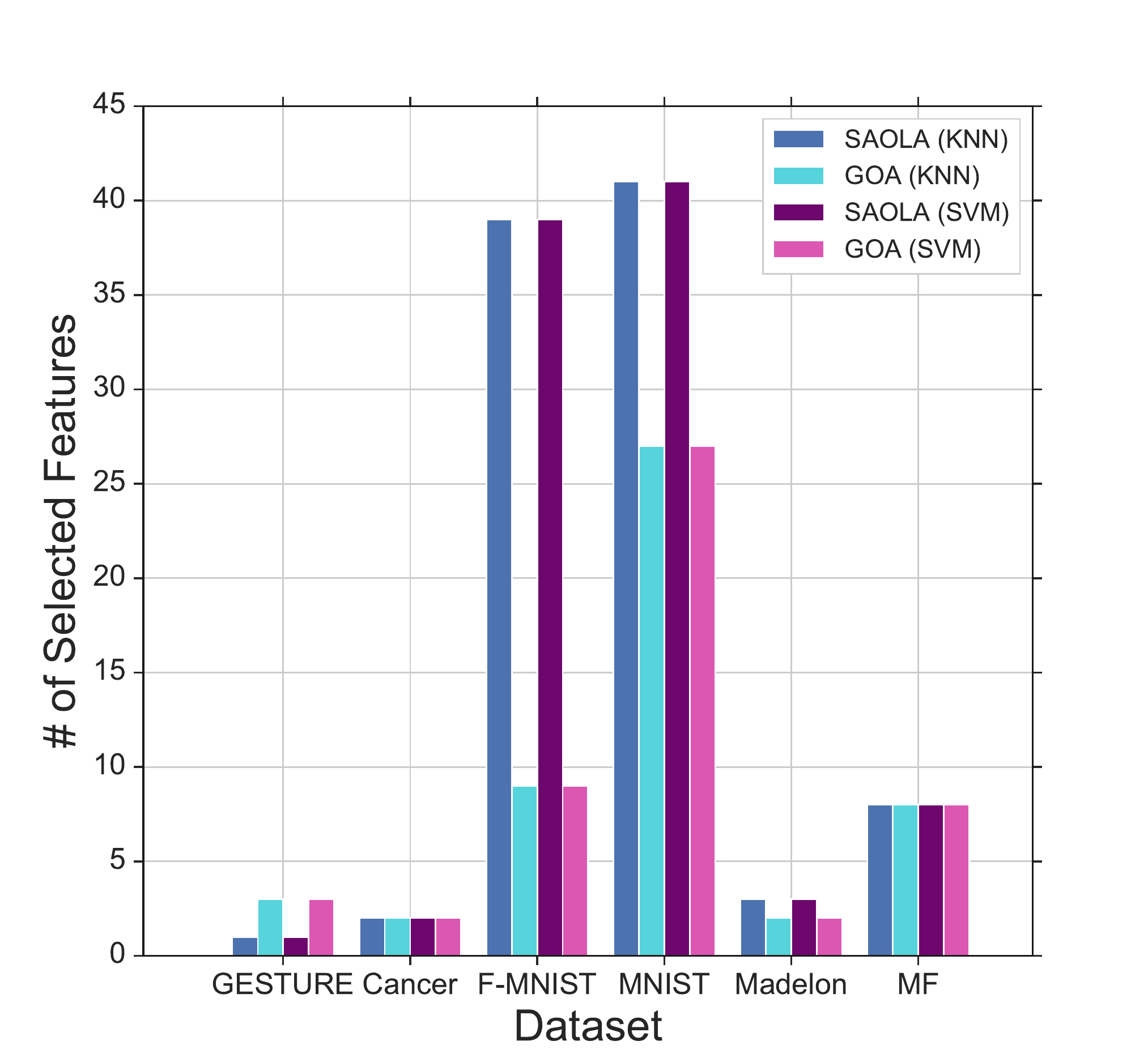}}    
      \caption{Our Geometric Online Adaptation method consistently outperforms Online Streaming Feature Selection baseline Alpha-Investing~\cite{Zhouetal2005}, Fast-OSFS~\cite{Wuetal.2013} and SAOLA~\cite{Yuetal2014}, while using less then or equal to the number of features across each baseline. This figure is explained in greater detail in our experimental section.}
      \label{figfeatures}
  \end{figure*}

 \subsection{OSFS - Fixed Sample}
 \label{osfs-fixed-sample}

 {\bf Simulation Results:} In the first case, we work with a 10-class classification problem where samples are of dimension $d=10$ and we vary the total number of samples per class from $100$ to $1000$. 
 In Fig.~\ref{fig:samplesize}, we observe that GOA consistently outperforms SAOLA and as the sample size increases, GOA's average accuracy improves across both SVM and KNN classifiers. 
 This is because the $\widehat{G}_\rho$ estimation algorithm becomes more accurate with increasing sample size. 
 However, when varying the dimensionality of features, we observe a slight dip in GOA's accuracy (Fig.~\ref{fig:dimsize}) since $\widehat{G}_\rho$ estimation becomes less accurate for larger dimensions. 
 Finally, Fig.~\ref{fig:classsize} shows that GOA is comparable if not slightly better than SAOLA for a variety of class sizes including $m=\{3,9,\ldots,20\}$, across both classifiers.

\begin{table*}
%\begin{table}[]
\caption {Relative average performance accuracy between GOA, Alpha-Investing, Fast-OSFS, and SAOLA on real-world datasets. GOA outperforms all baseline across most datasets and classifiers. Larger values  are better for performance while a corresponding small value is better for the number of features.}
\label{KNN1-RW}
\centering
\begin{tabular}{@{}lcc|cc|cc|cc|cc|cc@{}}
\toprule
\multirow{2}{*}{Dataset} & \multicolumn{2}{c}{Accuracy} & \multicolumn{2}{c}{Feats}  & \multicolumn{2}{c}{Accuracy} & \multicolumn{2}{c}{Feats} & \multicolumn{2}{c}{Accuracy} & \multicolumn{2}{c}{Feats} \\ 
 & \multicolumn{1}{c}{SAOLA} & \multicolumn{1}{c}{GOA} & \multicolumn{1}{c}{SAOLA} & \multicolumn{1}{c}{GOA} & \multicolumn{1}{c}{F-OSFS} & \multicolumn{1}{c}{GOA} & \multicolumn{1}{c}{F-OSFS} & \multicolumn{1}{c}{GOA} & \multicolumn{1}{c}{AI} & \multicolumn{1}{c}{GOA} & \multicolumn{1}{c}{AI} & \multicolumn{1}{c}{GOA} \\  \midrule 
                        & \multicolumn{12}{c}{KNN}                      \\ \midrule
GESTURE                 & 0.441 & \bf{0.846} & \bf{1} & 3 & 0.628 & \bf{0.846} & \bf{2} & 3 & 0.782 & \bf{0.846} & 14 & \bf{3} \\ 
WDBC                    & 0.751 & \bf{0.916} & 2 & 2 & 0.900 & \bf{0.923} & 4 & \bf{2} & \bf{0.938} & 0.930 & 17 & \bf{4} \\
FMNIST                  & 0.527 & \bf{0.775} & 39 & \bf{9} & 0.527 & \bf{0.775} & 39 & \bf{9} & 0.539 & \bf{0.775} & 43 & \bf{9} \\
MNIST                   & \bf{0.559} & 0.387 & 41 & \bf{27} & \bf{0.559} & 0.387 & 41 & \bf{27} & \bf{0.553} & 0.387 & 41 & \bf{27} \\ 
Madelon                 & 0.575 & \bf{0.583} & 3 & \bf{2} & 0.571 & \bf{0.583} & 3 & \bf{2} & \bf{0.702} & 0.637 & 5 & \bf{4} \\
MFEAT                   & 0.350 & \bf{0.839} & 8 & 8 & 0.406 & \bf{0.919} & 14 & 14 & 0.924 & \bf{0.970} & 149 & \bf{37}\\ \midrule 
                        & \multicolumn{12}{c}{SVM}  \\ \midrule
GESTURE                 & \bf{0.399} & 0.359 & \bf{1} & 3 & \bf{0.458} & 0.359 & \bf{2} & 3 & \bf{0.514} & 0.359 & 14 & \bf{3} \\  
WDBC                    & 0.812 & \bf{0.915} & 2 & 2 & 0.876 & \bf{0.940} & 4 & 4 & 0.944 & \bf{0.954} & 17 & \bf{5} \\
FMNIST                  & 0.393 & \bf{0.731} & 39 & \bf{9} & 0.393 & \bf{0.731} & 39 & \bf{9} & 0.405 & \bf{0.731} & 43 & \bf{9} \\
MNIST                   & \bf{0.543} & 0.326 & 41 & \bf{27} & \bf{0.543} & 0.326 & 41 & \bf{27} & \bf{0.543} & 0.326 & 41 & \bf{27} \\ 
Madelon                 & \bf{0.608} & 0.600 & 3 & \bf{2} & \bf{0.615} & 0.600 & 3 & \bf{2} & \bf{0.607} & 0.603 & 5 & 5 \\
MFEAT                   & 0.598 & \bf{0.798} & 8 & 8 & 0.797 & \bf{0.899} & 14 & 14 & 0.834 & \bf{0.966} & 149 & \bf{46}\\ \bottomrule
\end{tabular}
%\end{table}
\end{table*}

 {\bf Real-world Data Results:} To ground GOA's performance against available OSFS baselines we compare their performance on real-world datasets. 
 Table~\ref{KNN1-RW} summarizes the prediction accuracy of GOA against each of the baselines, Alpha-investing, Fast-OSFS, and SAOLA, individually, respectively. 
 We validate that GOA outperforms all the baselines in terms of prediction accuracy across both classifiers on most datasets while using an equivalent or less number features than the compared baseline.
  
Fig.~\ref{figfeatures} and Table~\ref{KNN1-RW} shows the number of selected features across all four algorithms. 
Alpha-investing selects, relatively, the most features but is still not able to compete with GOA. 
This is because Alpha-Investing never evaluates the redundancy of selected features and continues to accumulate features that could lead to many confounding factors. 
The advantage of using GOA is that it selects a very small number of features, oftentimes as low as 1, via the elimination of redundant features, while showcasing higher predictive performance. 
This means that GOA identifies the most informative features for each class variable and therefore is more accurate on the testing set. 
An interesting observation is that for the SVM classifier, on some datasets, GOA performs slightly worse than the baselines.
However, the difference in terms of accuracy is not large when compared to the difference in the number of features selected (for instance in GESTURE it is 14 vs 3 for Alpha-Investing).

\begin{table*}[!ht]
\caption{In average relative performance accuracy and number of features GOA outperforms extended SAOLA (X-S) on most real-world datasets for OSFS-SS. Larger values are better for performance while a corresponding small value is better for the number of features}
\label{streaming_results}
%\begin{table}[]
\centering
\begin{tabular}{@{}lcc|cc|cc|cc|cc|cc@{}}
\toprule
\multirow{2}{*}{Dataset} & \multicolumn{4}{c}{Scenario B\RNum{1}}    & \multicolumn{4}{c}{Scenario B\RNum{2}}    & \multicolumn{4}{c}{Scenario B\RNum{3}} \\
                        & \multicolumn{2}{c}{Accuracy} & \multicolumn{2}{c}{Feats.} & \multicolumn{2}{c}{Accuracy} & \multicolumn{2}{c}{Feats.}& \multicolumn{2}{c}{Accuracy} & \multicolumn{2}{c}{Feats.} \\ \midrule 
                        & \multicolumn{1}{c}{X-S} & \multicolumn{1}{c}{GOA} &  \multicolumn{1}{c}{X-S} & \multicolumn{1}{c}{GOA} & \multicolumn{1}{c}{X-S} & \multicolumn{1}{c}{GOA} & \multicolumn{1}{c}{X-S} & \multicolumn{1}{c}{GOA} & \multicolumn{1}{c}{X-S} & \multicolumn{1}{c}{GOA} & \multicolumn{1}{c}{X-S} & \multicolumn{1}{c}{GOA} \\ \midrule 
                        & \multicolumn{12}{c}{KNN}                      \\ \midrule
WDBC                    & 0.877 & \bf{0.881} & 3  & \bf{2} & 0.904 & \bf{0.917} & 3 & 3  & N.A. & N.A & - & -\\
MNIST                   & 0.535 & \bf{0.586} & 15 & \bf{5} & \bf{0.590} & 0.567 & 14 & \bf{5} & 0.505 &  \bf{0.549} & 11 & \bf{5}\\ 
FMNIST                  & 0.587 & \bf{0.671} & 9  & \bf{5} & 0.565 & \bf{0.651} & 9 &  \bf{6}  & 0.546 & \bf{0.666} & 8  & \bf{5}\\
GESTURE                 & 0.431 & \bf{0.639} & \bf{1}  & 2 & 0.490 & \bf{0.644} & \bf{1} & 3  & 0.410 & \bf{0.635} & \bf{1}  & 3\\ \midrule 
                        & \multicolumn{12}{c}{SVM}  \\ \midrule
WDBC                    & 0.568 & \bf{0.899} & 3  & \bf{2} & 0.761 & \bf{0.926} & 3 & \bf{3} & N.A. & N.A. & - & - \\
MNIST                   & 0.545 & \bf{0.580} & 15 & \bf{5} & 0.588 & \bf{0.606} & 14 &\bf{7} & 0.502 & \bf{0.574} & 8 & \bf{5}\\ 
FMNIST                  & 0.508 & \bf{0.617} & 9  & \bf{5} & 0.512 & \bf{0.586} & 9 & \bf{6} & 0.495 & \bf{0.656} & 9 & \bf{5}\\
GESTURE                 & \bf{0.414} & 0.378 & \bf{1}  & 2 & \bf{0.424} & 0.370 & \bf{1} & 3  &0.355 & \bf{0.359} & \bf{1} & 3 \\ \bottomrule
\end{tabular}
%\end{table}

\end{table*}

 %-----------------------------------------------------
 
 %---------------------
\subsection{OSFS - Streaming Sample}
GOA's full potential is harnessed when applied to real-world problems that involve streaming both samples and features.

{\bf Real-world Data Results:}
Table~\ref{streaming_results} clearly highlights the improvement in performance achieved when using GOA as compared to the extended SAOLA algorithm on real-world datasets.
Interestingly, basic intuition would suggest that as we stream smaller blocks of data, the performance of the algorithm should suffer since the estimator would become less accurate.
However, the features provided for the real-world datasets suggest that they are sufficient for the estimator to predict the underlying distribution well.\\
\noindent \textbf{Note:} Due to sample size and feature cardinality restrictions on the limited set of features in WDBC dataset, we only consider Scenarios B\RNum{1} \& B\RNum{2} while Scenario B\RNum{3} is not applied.

% Compared to OSFS, we observe that GOA alters its behaviour with variable or smaller size data blocks, specifically for deep features.
% In results from Scenario B\RNum{1}, the expected behaviour of GOA selecting lesser than or equal to the number of features in SAOLA is matched.
% However, in scenarios B\RNum{2} and B\RNum{3}, GOA defaults to selecting a slightly larger number of features as opposed to modified SAOLA.
% This is clearly illustrated from the results for CIFAR-10, where the number of selected features for GOA are 4, 8 and 18 while SAOLA defaults to a lower number of features, 4, 2 and 4, while maintaining lower performance.
% A similar variation is observable in results for the KTH dataset.\\

 %-----------------------------------------------

 %-----------------------------------------------
\subsection{Visualization}
In this section we visualize the PCA components selected by GOA (Fig.~\ref{OSFS-SS-GOA}) and extended SAOLA (Fig.~\ref{OSFS-SS-SAOLA}) in Scenario B\RNum{2} from the Fashion MNIST dataset.
There are common four principal components between the feature subsets selected by each algorithm, 1, 2, 5 and 23.
Interestingly, GOA's performance surpasses extended SAOLA even though the number of highly weighted principal components, as well as the number of selected principal components, is larger in extended SAOLA.
This interesting outcome further supplements our claim that the features selected by GOA, under the comparison criterion used in the experiments, are more informative to the class label than those selected by SAOLA.

 \begin{figure}[h!]
       \centering
      \subfloat[\rm \bf F1]{\includegraphics[width=0.15\columnwidth]{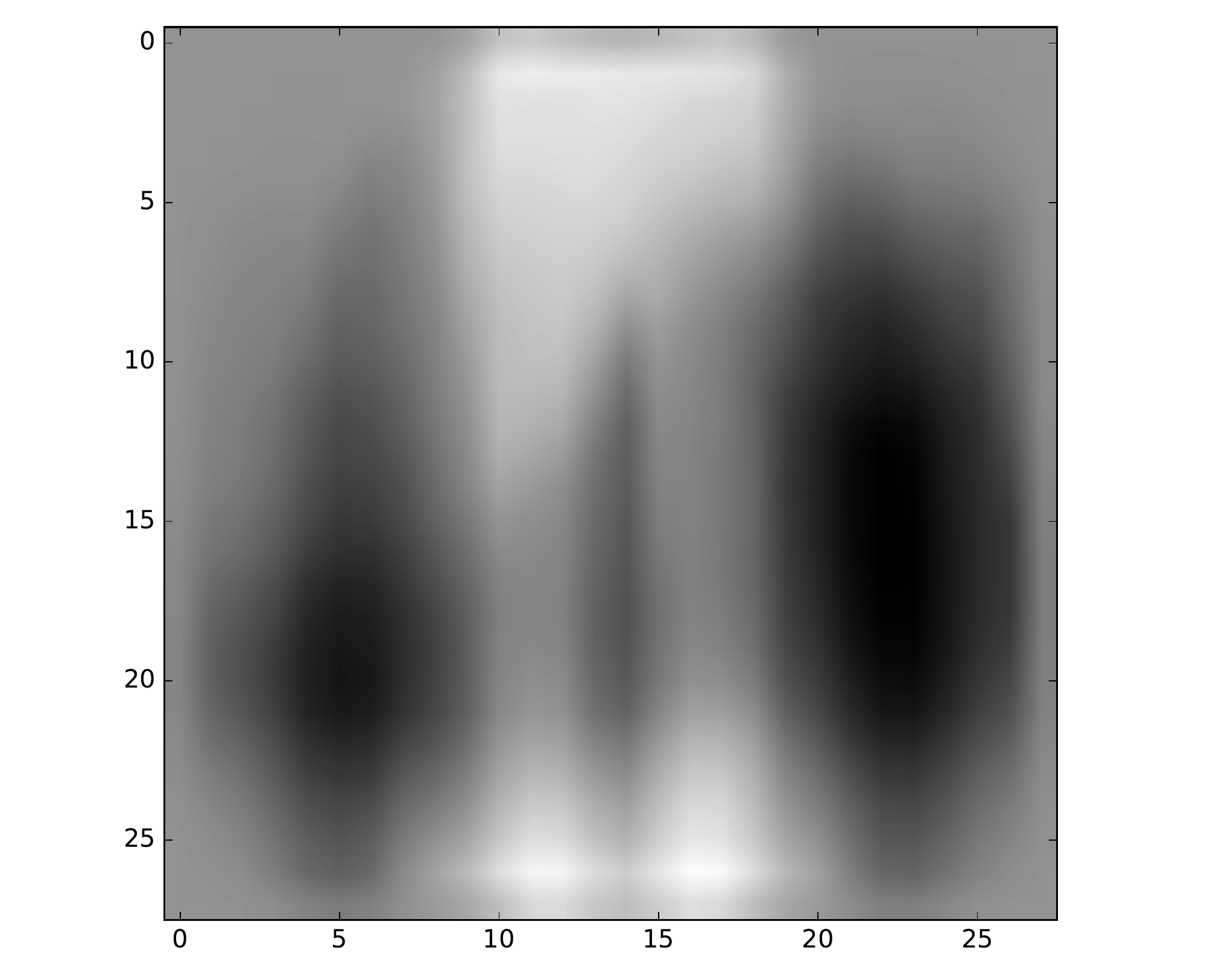}}    
      \subfloat[\rm \bf F2]{\includegraphics[width=0.15\columnwidth]{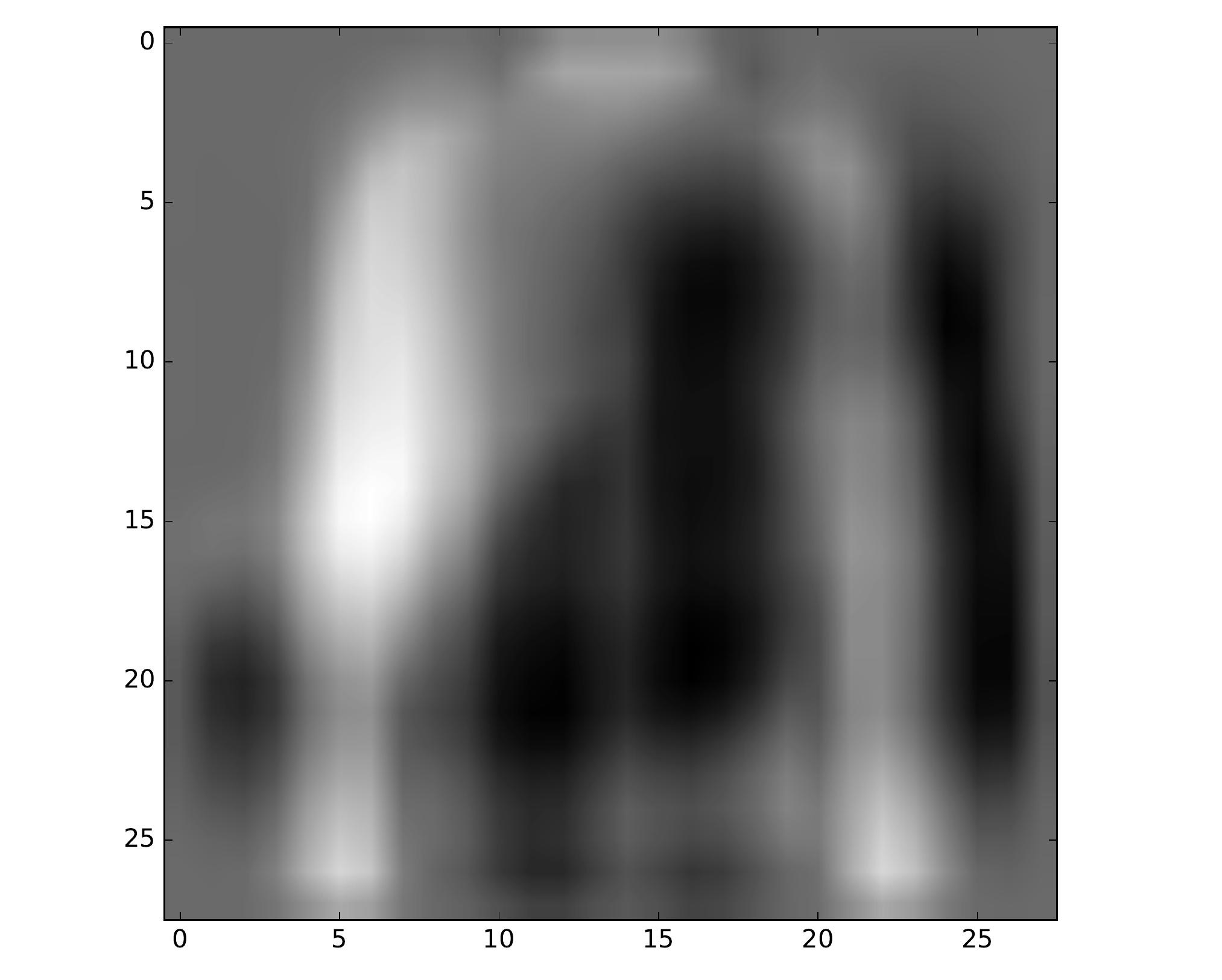}}    
      \subfloat[\rm \bf F5]{\includegraphics[width=0.15\columnwidth]{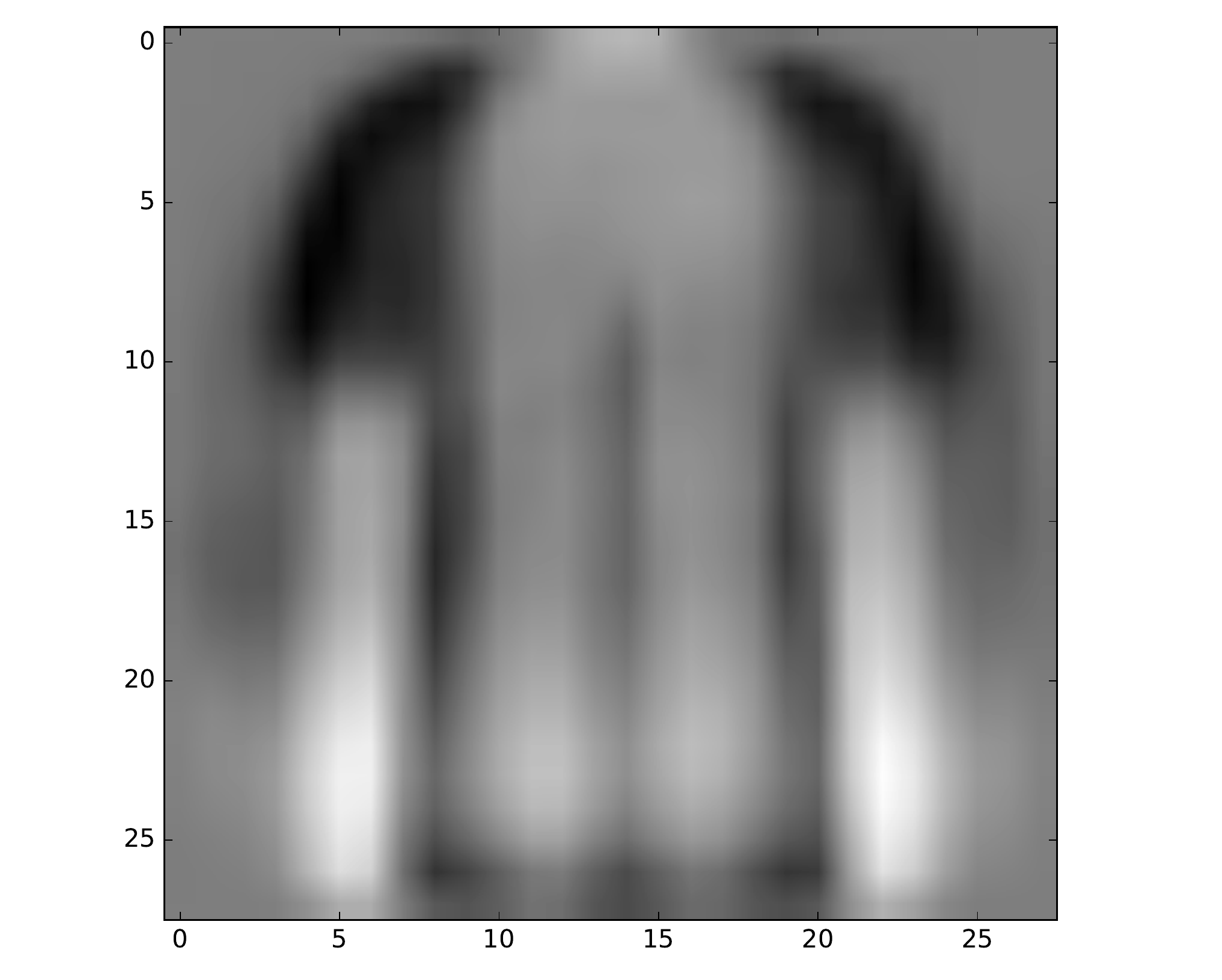}}    
      \subfloat[\rm \bf F7]{\includegraphics[width=0.15\columnwidth]{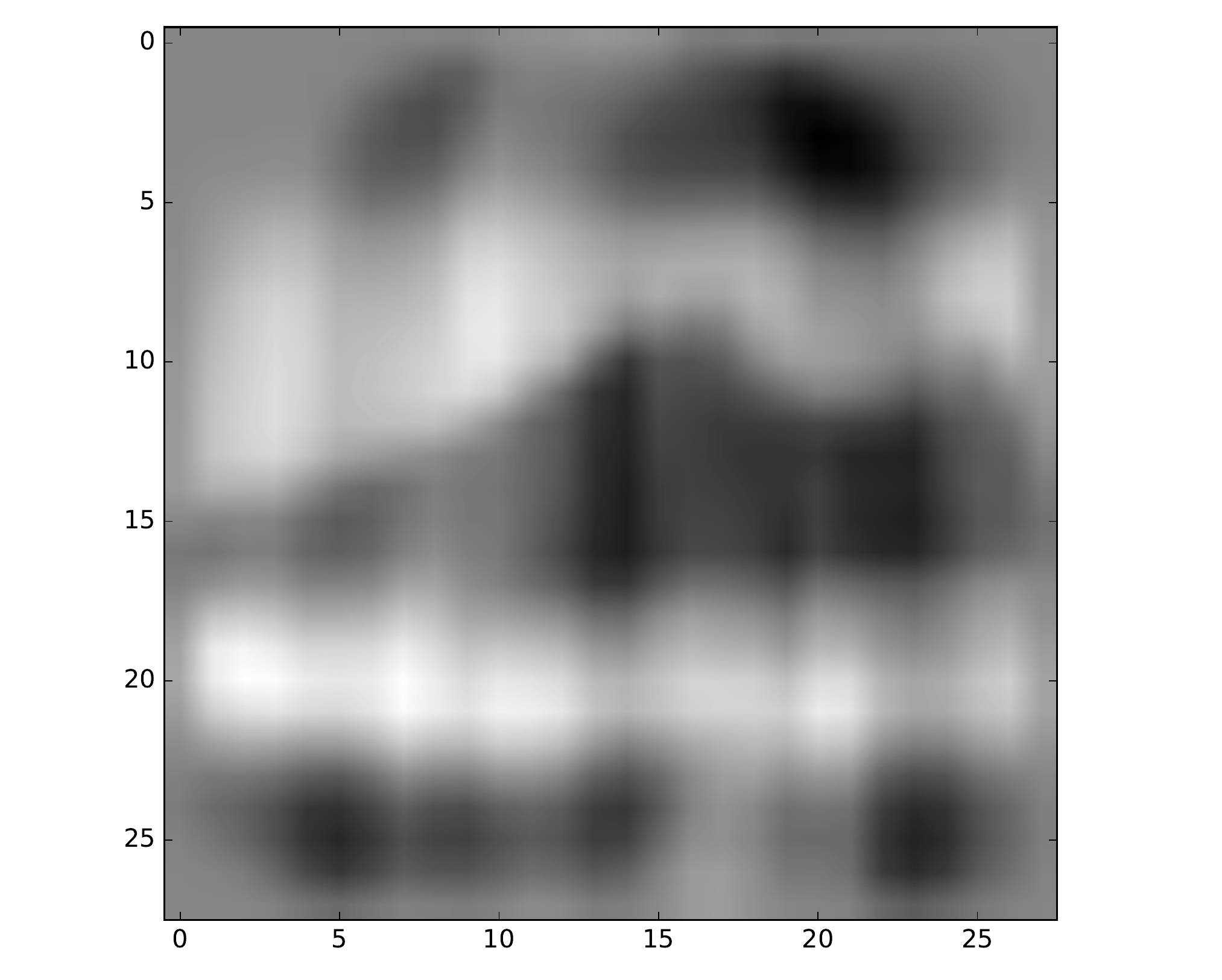}}    
      \subfloat[\rm \bf F23]{\includegraphics[width=0.15\columnwidth]{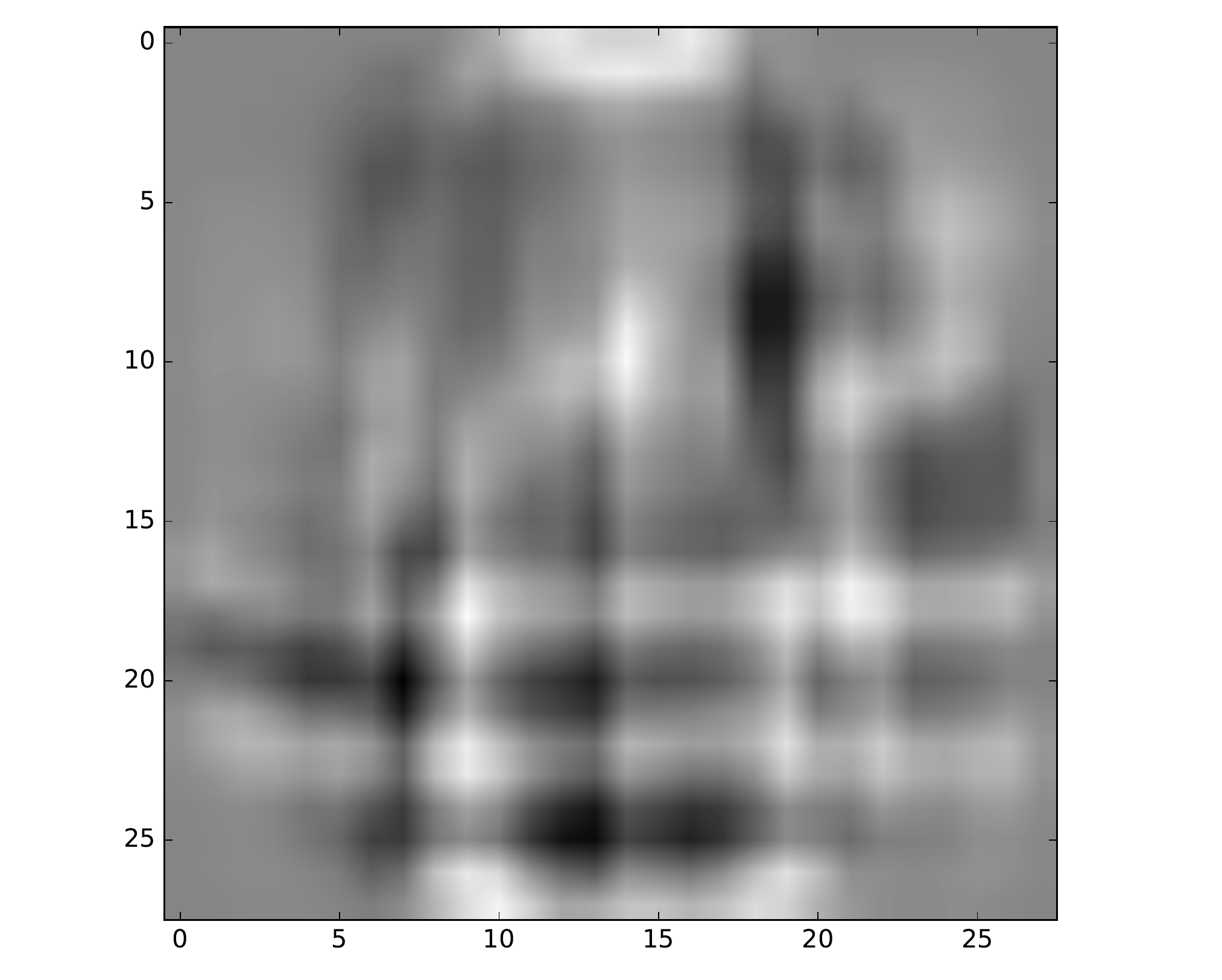}}    
      \subfloat[\rm \bf F34]{\includegraphics[width=0.15\columnwidth]{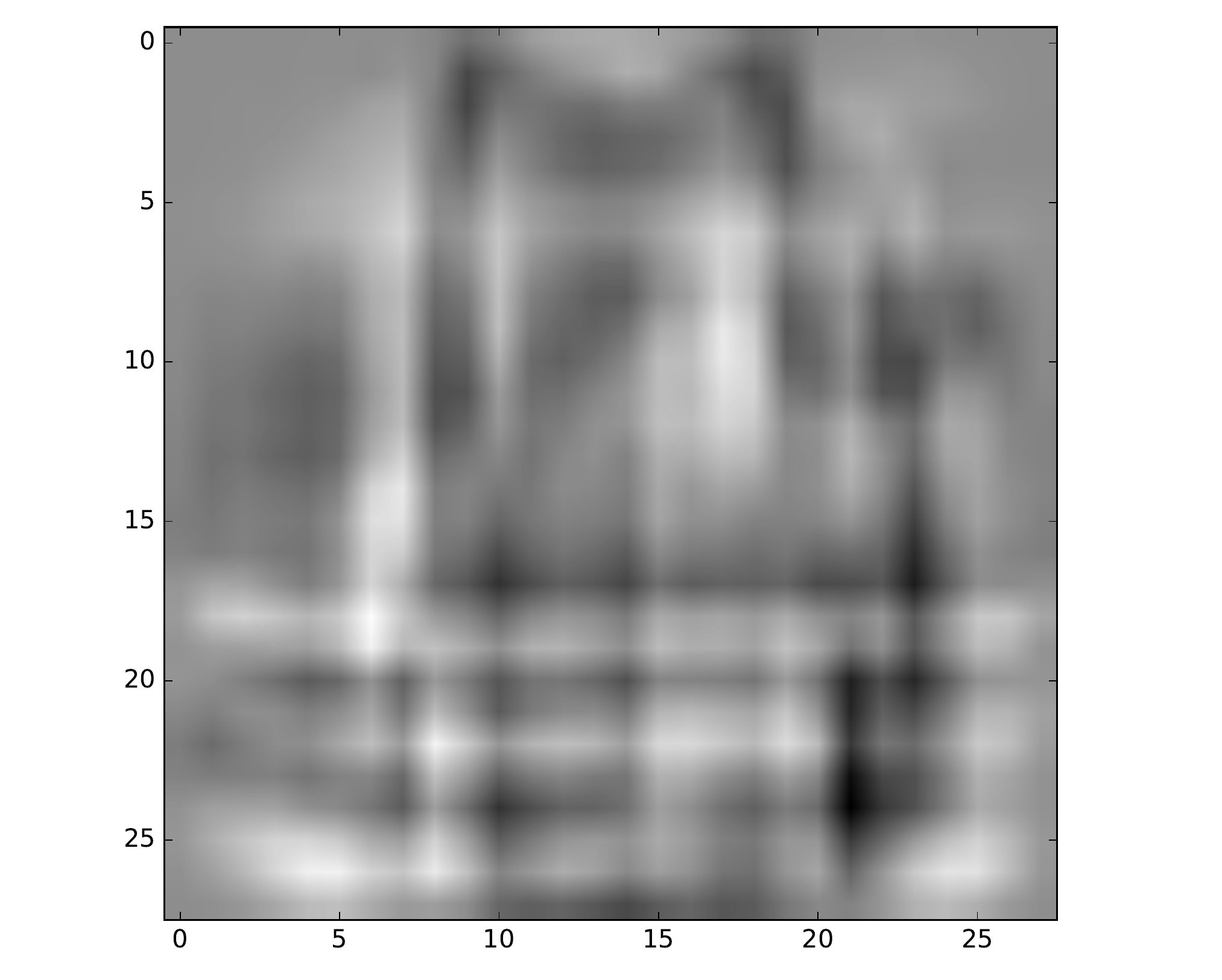}}    
       \caption{Illustration of the 6 principal components selected by GOA (Alg.~\ref{algorithm0}) from Fashion MNIST data in Scenario B\RNum{2}. The numbers underneath the figure represent the principal components, in order.}
       \label{OSFS-SS-GOA}
   \end{figure}
  \begin{figure}[h!]
      \centering
      \subfloat[\rm \bf F1]{\includegraphics[width=0.15\columnwidth]{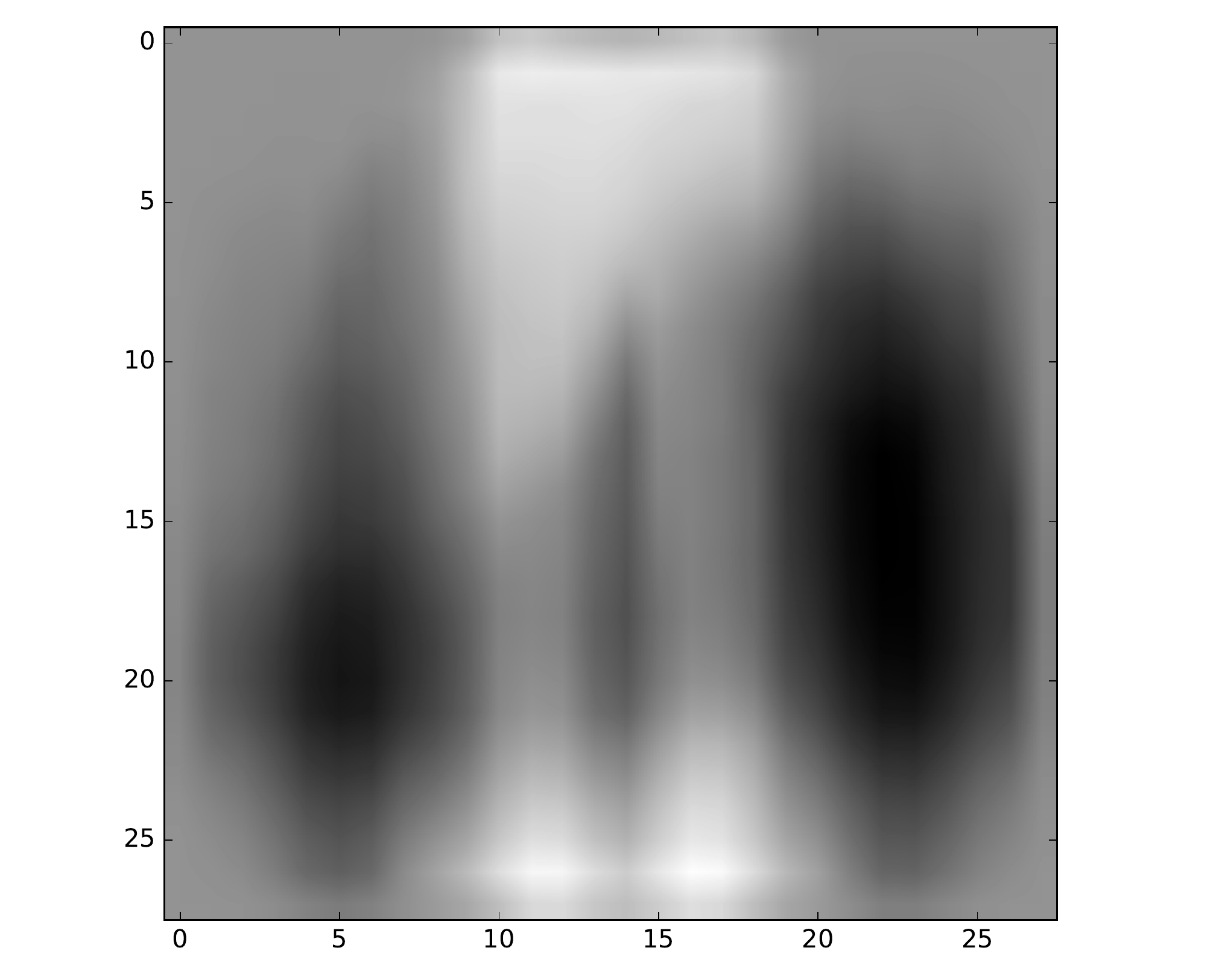}}    
      \subfloat[\rm \bf F2]{\includegraphics[width=0.15\columnwidth]{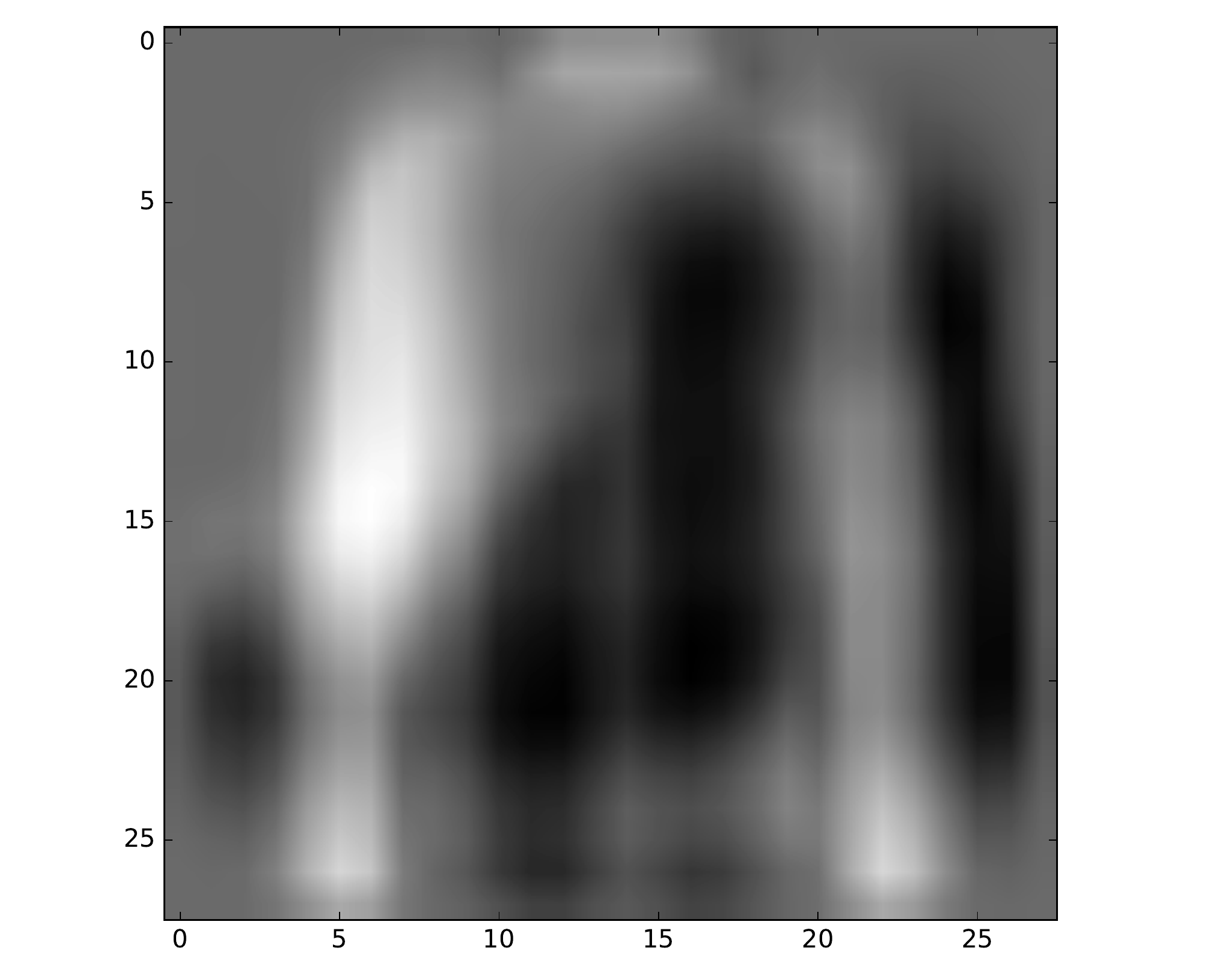}}    
      \subfloat[\rm \bf F3]{\includegraphics[width=0.15\columnwidth]{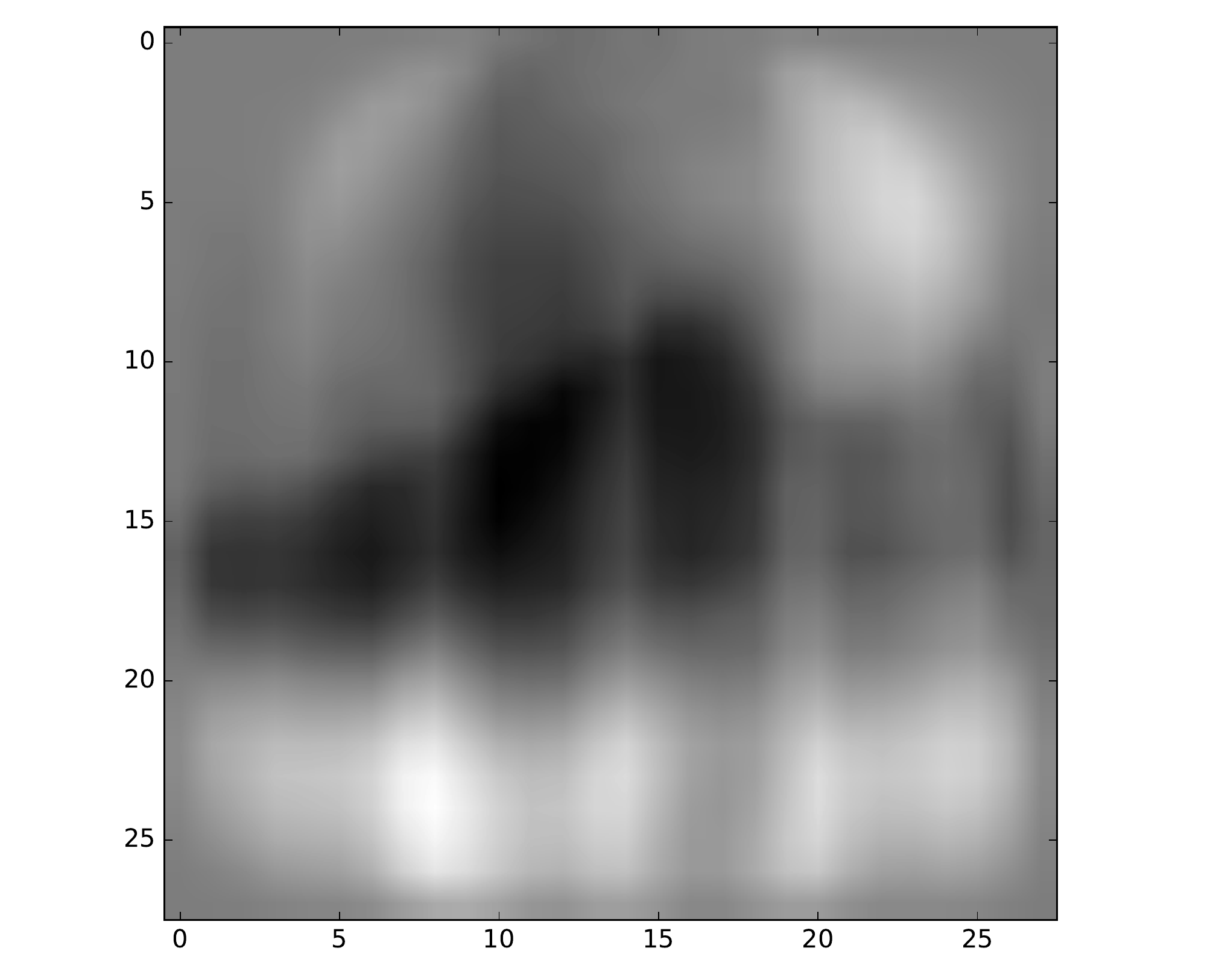}}    
      \subfloat[\rm \bf F5]{\includegraphics[width=0.15\columnwidth]{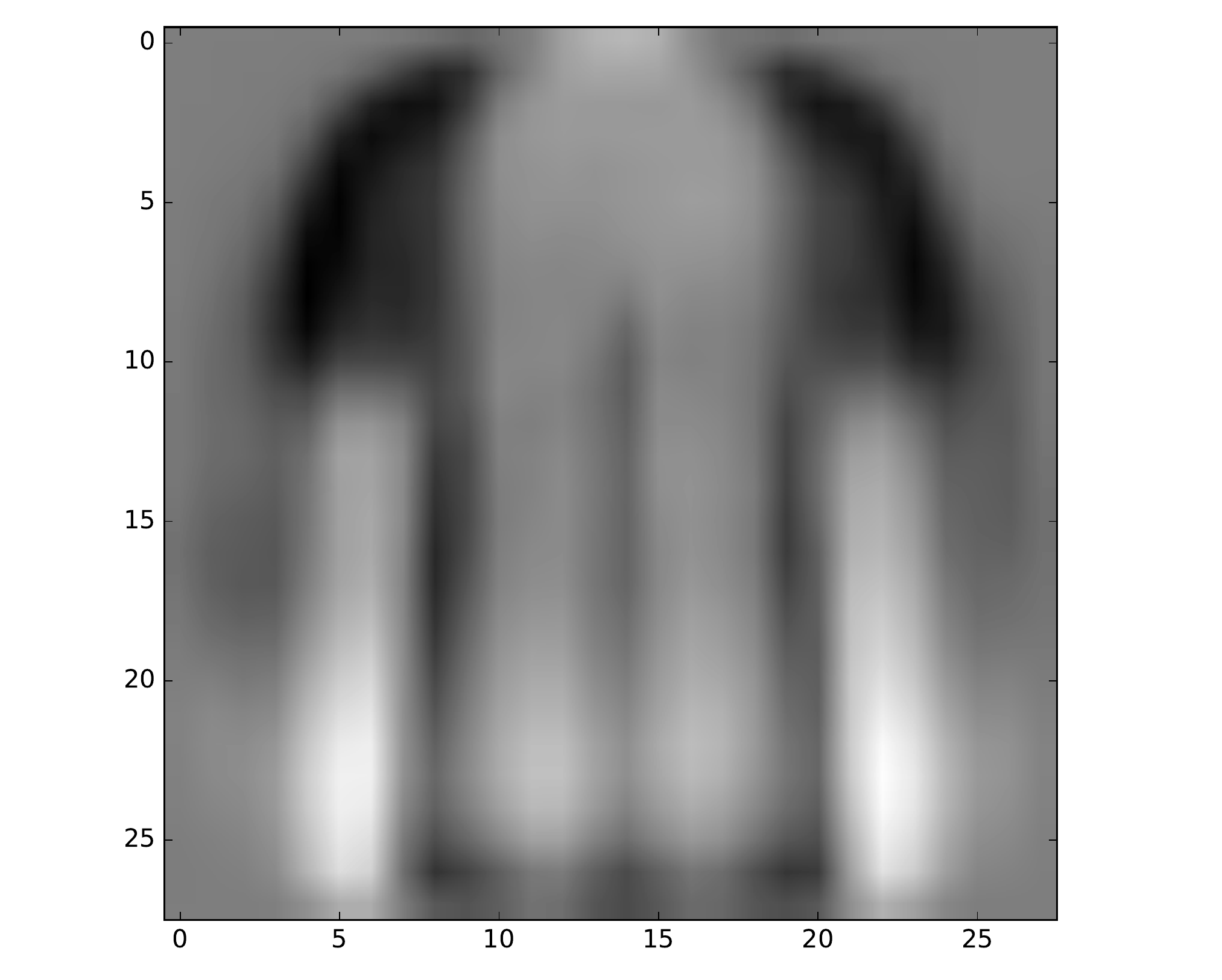}}    
      \subfloat[\rm \bf F6]{\includegraphics[width=0.15\columnwidth]{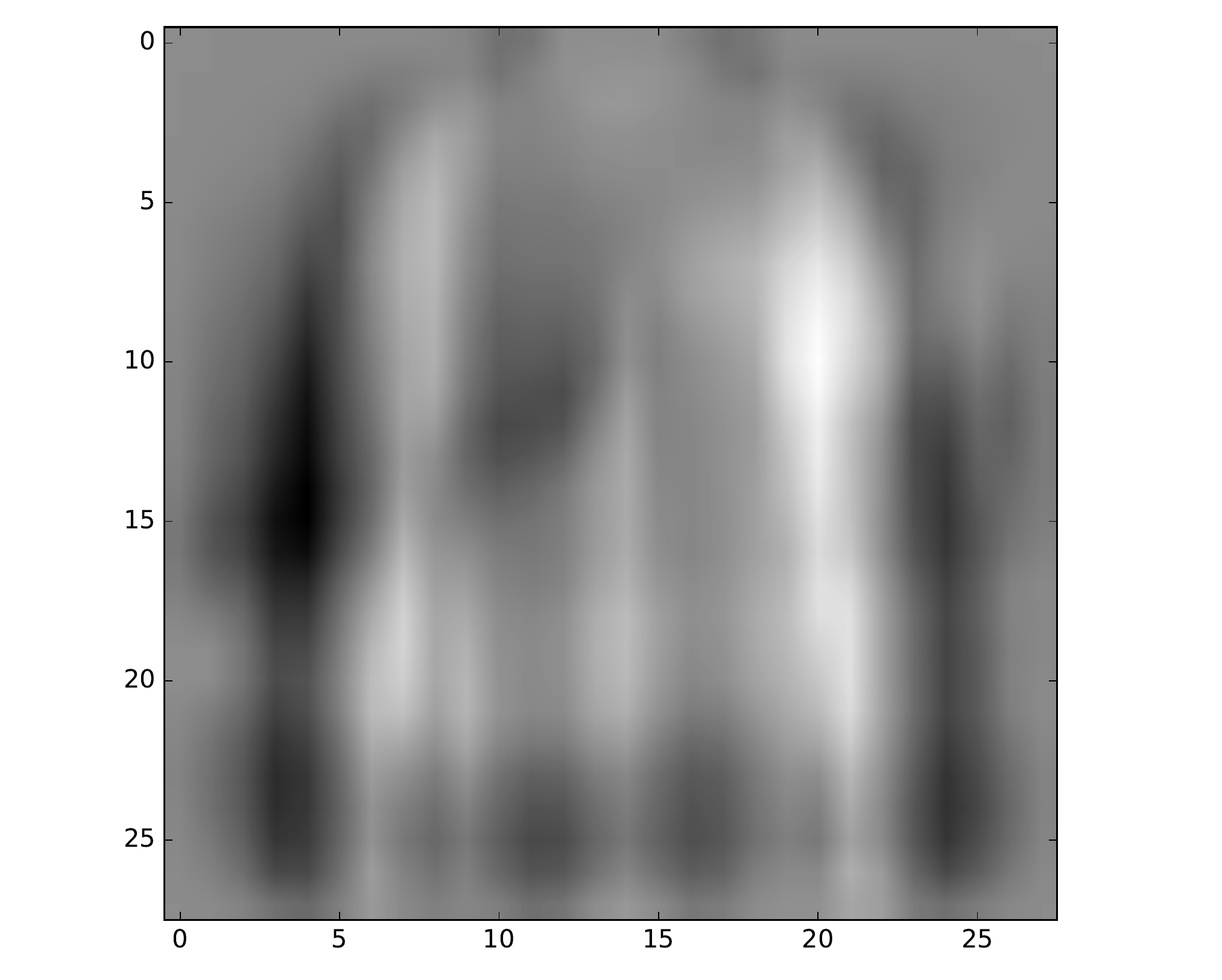}} \\    
      \subfloat[\rm \bf F9]{\includegraphics[width=0.15\columnwidth]{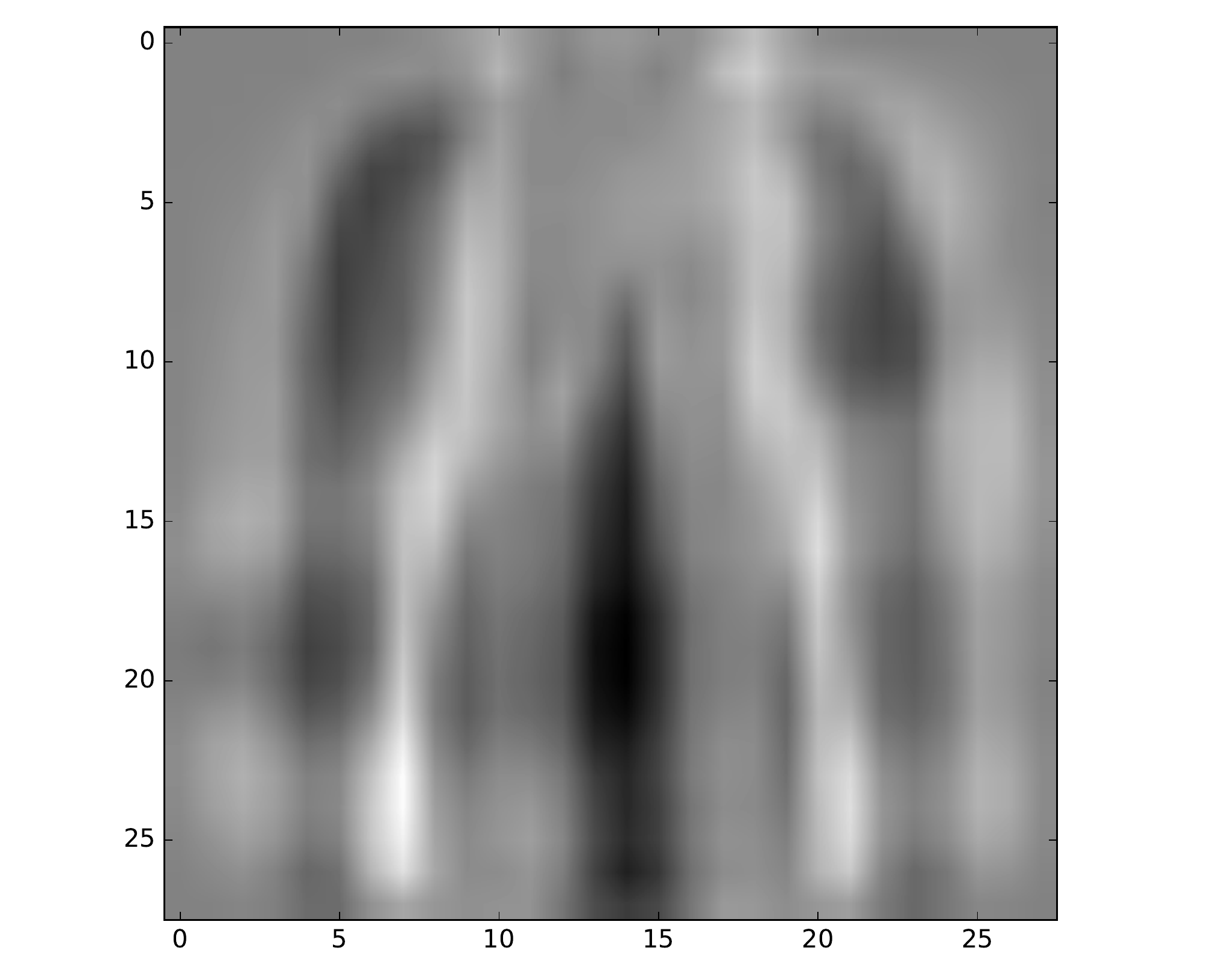}}    
      \subfloat[\rm \bf F11]{\includegraphics[width=0.15\columnwidth]{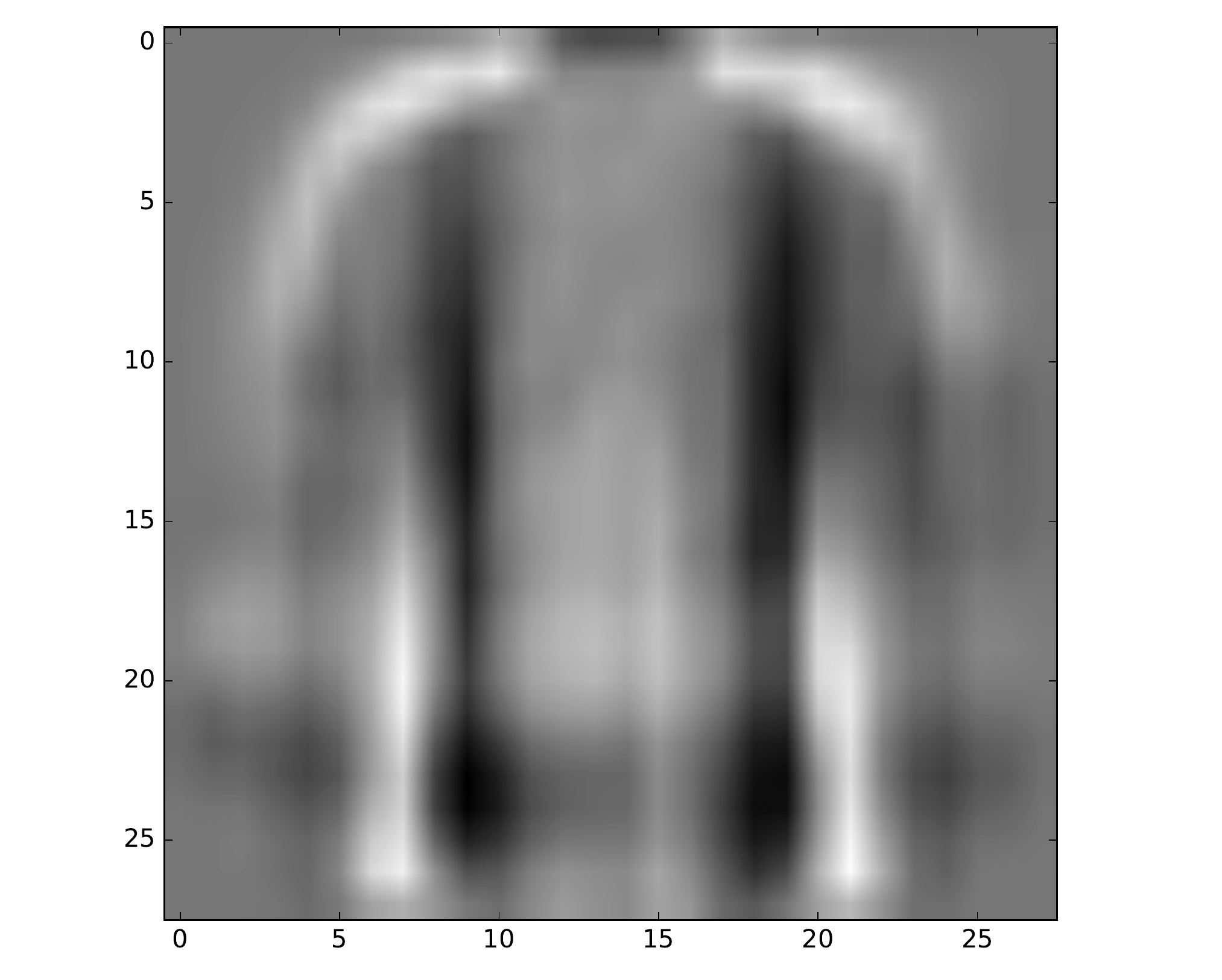}}    
      \subfloat[\rm \bf F18]{\includegraphics[width=0.15\columnwidth]{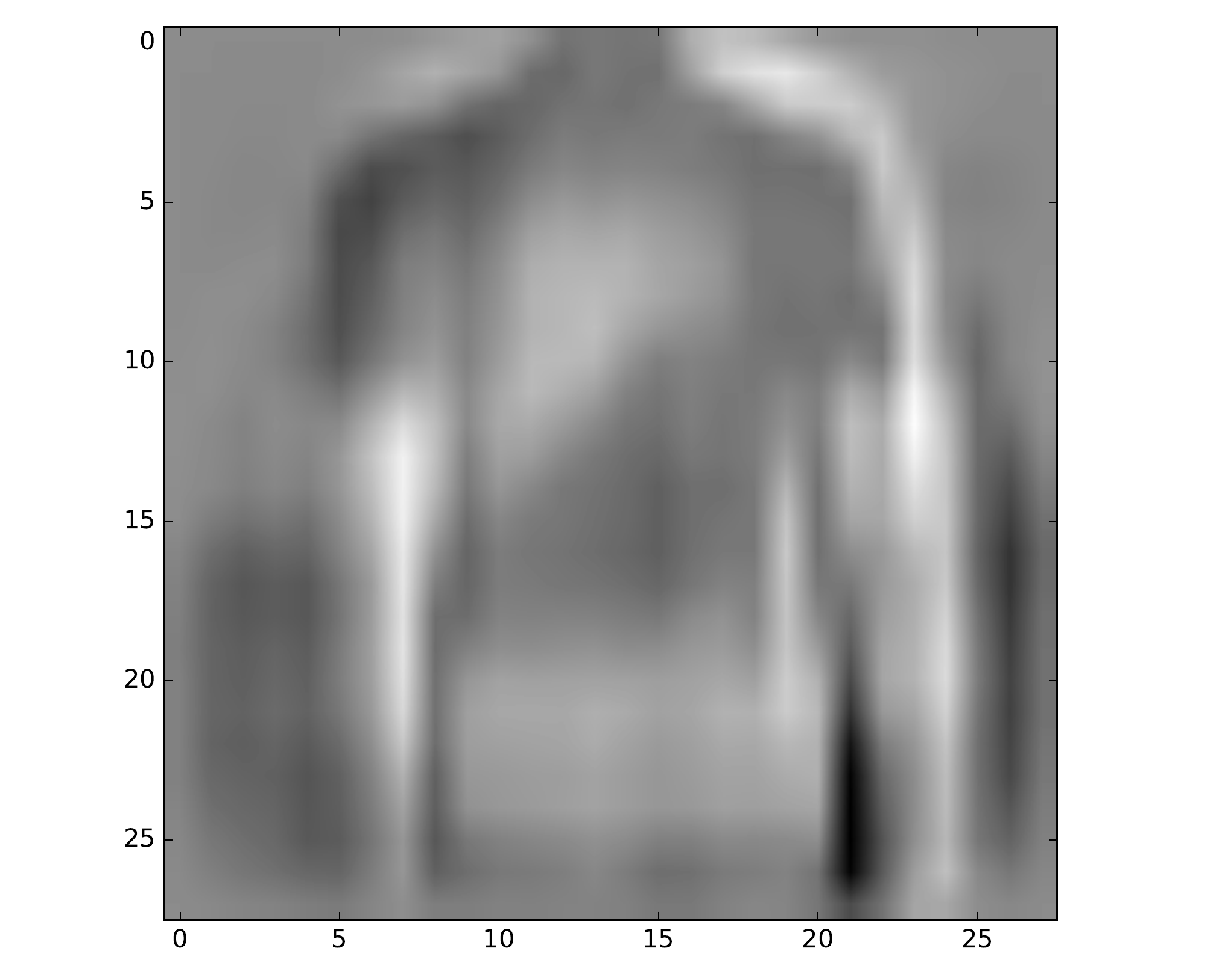}}    
      \subfloat[\rm \bf F20]{\includegraphics[width=0.15\columnwidth]{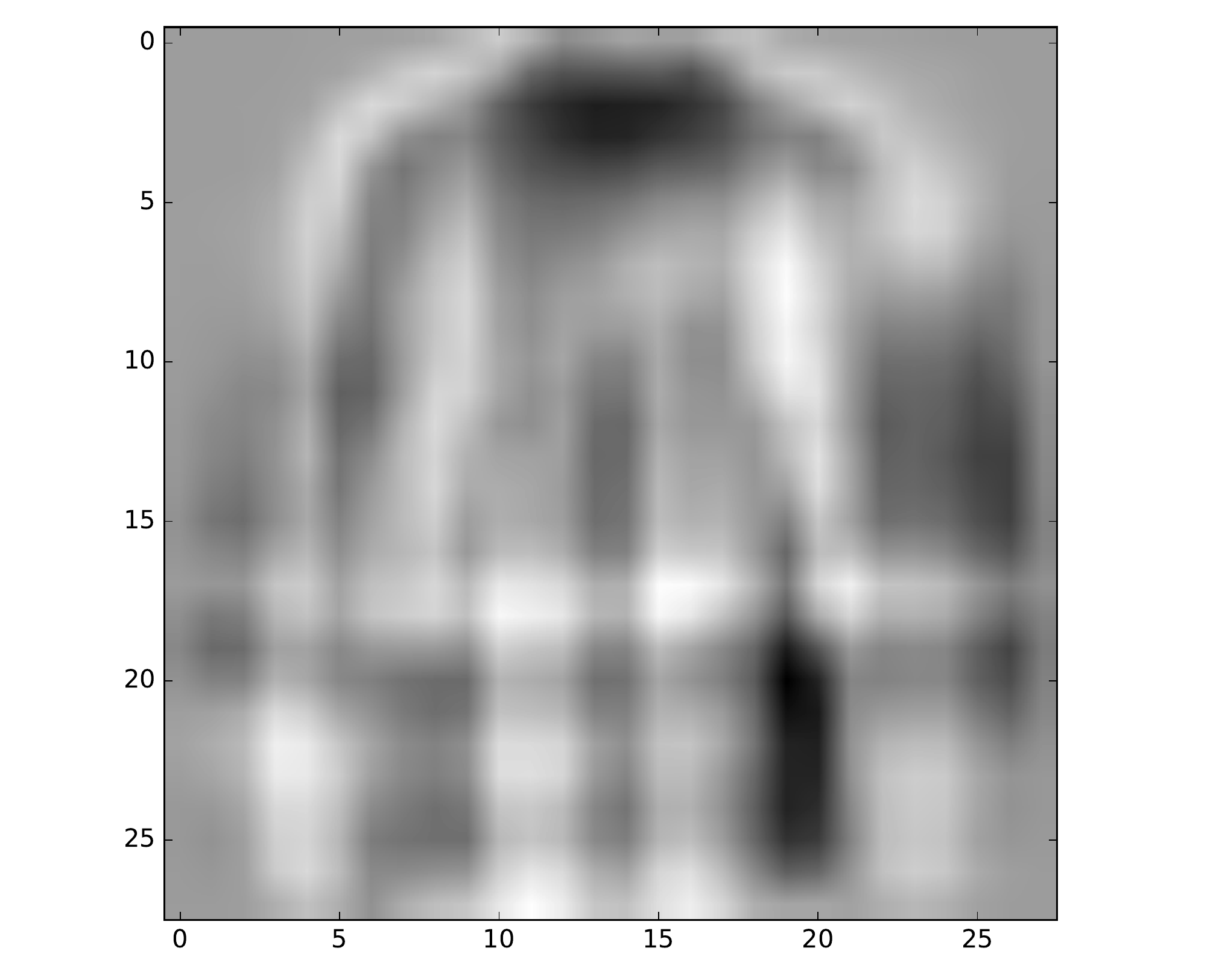}}    
      \subfloat[\rm \bf F23]{\includegraphics[width=0.15\columnwidth]{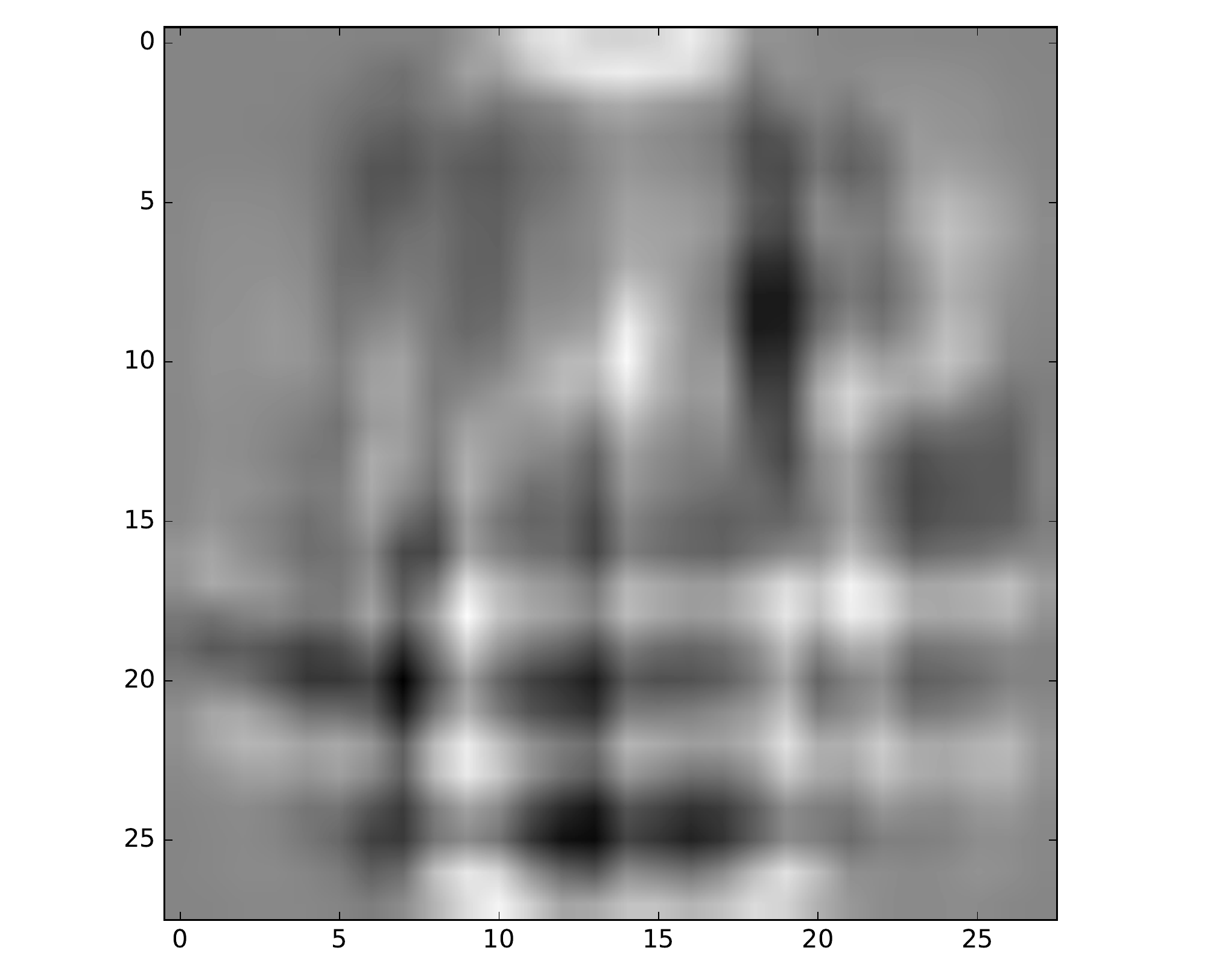}}    
  \caption{Illustration of the 10 principal components selected by extended SAOLA from Fashion MNIST data in Scenario B\RNum{2}. The numbers underneath the figure represent the principal component order. Extended SAOLA selects a larger number of components, both in total and ratio of highly valued principal components, and still performs worse than GOA.}
      \label{OSFS-SS-SAOLA}
  \end{figure}

\section{Deep Features Application}
 \begin{table*}[!ht]
%\begin{table}[]
\caption{In relative average performance accuracy GOA outperforms extended SAOLA on all deep feature datasets for OSFS-SS. In scenarios B\RNum{2} and B\RNum{3}, GOA defaults to selecting a larger number of features due to uncertainty in the underlying distribution. Parallel to this observation, the number of features selected in B\RNum{3} is lower than B\RNum{2}.}
\label{deep_streaming_results}
\centering
\begin{tabular}{@{}lcc|cc|cc|cc|cc|cc@{}}
\toprule
\multirow{2}{*}{Dataset} & \multicolumn{4}{c}{Scenario B\RNum{1}}    & \multicolumn{4}{c}{Scenario B\RNum{2}}    & \multicolumn{4}{c}{Scenario B\RNum{3}} \\
                        & \multicolumn{2}{c}{Accuracy} & \multicolumn{2}{c}{Feats.} & \multicolumn{2}{c}{Accuracy} & \multicolumn{2}{c}{Feats.}& \multicolumn{2}{c}{Accuracy} & \multicolumn{2}{c}{Feats.} \\ \midrule 
                        & \multicolumn{1}{c}{X-S} & \multicolumn{1}{c}{GOA} &  \multicolumn{1}{c}{X-S} & \multicolumn{1}{c}{GOA} & \multicolumn{1}{c}{X-S} & \multicolumn{1}{c}{GOA} & \multicolumn{1}{c}{X-S} & \multicolumn{1}{c}{GOA} & \multicolumn{1}{c}{X-S} & \multicolumn{1}{c}{GOA} & \multicolumn{1}{c}{X-S} & \multicolumn{1}{c}{GOA} \\ \midrule 
                        & \multicolumn{12}{c}{KNN}  \\ \midrule
% \multirow{2}{*}{Dataset} & \multicolumn{2}{c}{Scenario BI}    & \multicolumn{2}{c}{Scenario BII}    & \multicolumn{2}{c}{Scenario BIII} \\
%                         & \multicolumn{6}{c}{KNN}                      \\ \midrule
%                         & $\Delta$ Acc. & $\Delta$ Feats.& $\Delta $Acc. & $\Delta $Feats.& $\Delta $Acc. & $\Delta $Feats. \\ \midrule 
CIFAR-10 - AlexNet        & 0.123 & \bf{0.203} & 7 & \bf{6} & 0.132 & \bf{0.263} & \bf{4} & 49 & 0.109 & \bf{0.255} & \bf{6} & 20 \\
CIFAR-10 - VGG16          & 0.199 & \bf{0.280} & 8 & \bf{7} & 0.154 & \bf{0.332} & \bf{2} & 44 & 0.173 & \bf{0.293} & \bf{3} & 20 \\ 
CIFAR-10 - ResNet18       & 0.161 & \bf{0.277} & 9 & \bf{5} & 0.136 & \bf{0.216} & \bf{7} & 18 & 0.100 & \bf{0.324} & \bf{6} & 21 \\
CIFAR-10 - ResNet34       & 0.125 & \bf{0.320} & 7 & \bf{5} & 0.143 & \bf{0.287} & \bf{3} & 25 & 0.117 & \bf{0.284} & \bf{6} & 18 \\ \midrule 
                        & \multicolumn{12}{c}{SVM}  \\ \midrule
CIFAR-10 - AlexNet        & 0.139 & \bf{0.290} & 8 & \bf{7} & 0.134 & \bf{0.226} & \bf{4} & 46 & 0.108 & \bf{0.316} & \bf{5} & 19 \\
CIFAR-10 - VGG16          & 0.183 & \bf{0.359} & 7 & 7      & 0.228 & \bf{0.368} & \bf{3} & 49 & 0.215 & \bf{0.328} & \bf{6} & 20 \\ 
CIFAR-10 - ResNet18       & 0.153 & \bf{0.338} & 6 & 6      & 0.159 & \bf{0.278} & \bf{4} & 33 & 0.121 & \bf{0.210} & \bf{4} & 16 \\
CIFAR-10 - ResNet34       & 0.177 & \bf{0.323} & 8 & \bf{6} & 0.207 & \bf{0.447} & \bf{5} & 18 & 0.100 & \bf{0.312} & \bf{4} & 19 \\ \bottomrule 
\end{tabular}
%\end{table}

\end{table*}
 
As an extension of the various streaming feature selection scenarios presented in this work, we apply GOA to features obtained by processing CIFAR-10 through a variety of deep neural networks.
They include commonly used deep networks such as AlexNet, VGG16, ResNet18 and ResNet34.
We note that we only apply simple preprocessing steps such as resizing and mean subtraction on the inputs to each network while the features themselves are PCA scaled versions of the outcomes from the second last layer of each network.

 {\bf Deep Feature Results:}
 From Table~\ref{deep_streaming_results}, we clearly observe that GOA outperforms the extended SAOLA algorithm on deep features, in addition to the real-world datasets.
 Results from scenario B\RNum{1} mirror those from both the simulation as well as real-world datasets with regard to improved mean recognition accuracy while selecting equivalent or lesser number of features than extended SAOLA.
 However, in scenarios B\RNum{2} and B\RNum{3}, we observe that GOA selects an extremely large number of features. 
 We hypothesize that GOA defaults to such a larger number of features when compared to the poorly performing extended version of SAOLA because it is tuned to maintain a high level of performance. 
 With large variations in the number of samples observed by GOA in scenarios B\RNum{2} and B\RNum{3}, the MI estimator's uncertainty on the underlying distribution of the data is large and reflected in the additional features accumulated by GOA.
 In supplement our hypothesis with the observation that the relative difference in the number of features selected is lower in scenario B\RNum{3} than B\RNum{2}, where a more fixed albeit small number of samples are observed.

\section{Conclusion}
In this paper, we challenge standard OSFS settings by introducing an online streaming feature selection setup where both features and samples are streamed simultaneously. 
We consider this a natural extension to OSFS and closer approximation to real-world problems.
Further, we propose GOA as a method applicable in both settings and show that it consistently outperforms relevant baselines while maintaining equivalent or smaller subsets of features across a variety of datasets.
Finally, when characterizing the behaviour of OSFS-SS algorithms on CNN-based features, we identify that accounting for evolving data distributions and the drift in feature importance cause by such distributions is critical to developing better methods for OSFS-SS in the future.
Apart from this, we plan to extend our method to real-time settings on high-dimensional datasets.

%\begin{acks}
%NEED TO EDIT THIS
%\end{acks}

\bibliographystyle{IEEEtran}
\bibliography{sample-base}

\newpage~\newpage

%\appendix
\section*{Supplementary Material}
%\subsection{Proof of Lemma \ref{lemma.1}:}
%Salimeh is working on this! 
\section{Further Discussion on Conjecture}
Let us restate the conjecture here once again:

 Given the current feature subset $S^*_{t_{i-1}}$ at time $t_{i-1}$ and a new feature $F_i$ at time $t_i$, if $\exists X,Z\in S^*_{t_{i-1}}$, $X\neq Z$ such that $I(F_i;C|Z)=0$ and $I(F_i;C|X,Z)=0$ then we have $I(F_i;Z|C)\geq I(X;Z|C)$.
\\
We explain this conjecture for when class variable $C$ takes only one label i.e. $C=i$ because conditional MI between two features given class variable $C$ is a weighted MI between features where weights are class prior probabilities. 
Now assume that for given features $X$ and $Z$, new incoming feature $F_i$ and class variable $C$ are independent. 
In addition, assume that $F_i$ and $C$ are independent given $Z$. 
This means that information provided by $Z$ causes $F_i$ to be less informative with respect to $C$ than the information provided by feature $X$. 
Another way to state this would be, $F_i$ is more dependent on $Z$ than $X$ given class $i$. Proving this conjecture logically is considered as a future work. 

%------------------------
\section{Further Discussion on Experiments}

\subsection{Speeding up the GOA algorithm}
The GOA method computes pairwise feature relevances using the CGD measure~\cite{ICASSPsalimeh2019} ($\widehat{G}_{\rho}$). 
Pairwise measures are repeatedly recomputed and require the computation of a Euclidian Minimum Spanning tree (EMST) over two dimensional points \cite{sabuncu2005gradient,sabuncu2005graph,sabuncu2008using} at every step. 
To speed up the algorithm, we pre-compute this measure for all feature pairs and store it in a data matrix that is repeatedly accessed during the execution of the algorithm. 
This pre-computation step significantly speeds up the algorithm. 
Usually EMSTs are not guaranteed to be unique and thereby the relevance between two features may change as a function of the instantiated tree. 
However, when using this pre-computation step, pairwise feature relevances are  fixed. 
This step does not appear to affect the algorithm's final performance significantly.

\subsection{Discussion on OSFS-Streaming Samples}
The streaming setup used in both OSFS and OSFS-SS experiments are illustrated in Fig.~\ref{fig.Scenarios}.
Scenario A (OSFS) only requires streaming features hence the temporal window takes the value of a specific feature across all the samples in the data matrix.
In all three scenarios in B temporal windows take non-overlapping subsets of samples and features.
An important constraint when selecting datasets for OSFS-SS is that each temporal window must contain sufficient samples to cover at least one sample per label in the dataset.
Hence, for a balanced dataset the ratio of total number of samples divided by the total number of features must be greater than the total number of unique labels. 
Even in scenario B\RNum{3}, for uneven temporal window sizes, the minimum number of samples in a window must cover at least one sample for every label in the dataset.
This constraint is necessary to ensure that the class label variable, $C$, is fixed and every iteration in Alg.~\ref{algorithm0} remains balanced.
Thus, in certain cases like KTH or WDBC for scenario B\RNum{3}, which requires multiple passes over a given feature thereby requiring a large amount of samples compared to the number of labels/features, the setting is not applicable. 

%-----------------------------------------------
 \subsection{SAOLA for Streaming Samples}
 
  \begin{algorithm}[h!]
  \begin{algorithmic}[1]
  \renewcommand{\algorithmicrequire}{\textbf{Input:}}
  \renewcommand{\algorithmicensure}{\textbf{Output:}}
  \REQUIRE $F_i$: predictive feature; $C$: the class labels, $0\leq\delta\leq 1$: a relevance threshold; \\$S^*_{t_{i-1}}$: the selected feature set at time $t_{i-1}$;
  $S^*_{t_{i}}$: the selected feature set at time $t_{i}$;
  $n_{t_i}$: the number of sample at time $t_i$;
 % $\{\bx_1,\bx_2,\ldots,\bx_{n_{t_i}}\}$: the observed streaming sample at time $t_i$;
  $\mathcal{A}_{t_i}$: the observed streaming sample matrix with feature set $S^*_{t_i-1}\cup F_i$ at time $t_i$;
   \vspace{0.2cm}
   \STATE {\bf Repeat}
   \STATE Get a new sample  $\mathcal{A}_{t_i}$ at time $t_i$, which includes a new feature $F_i$;
   \STATE \quad{\bf for} features $Y\in S^*_{t_{i-1}}$ {\bf compute}
   \STATE \quad \quad $\widehat{I}(F_i;C)$, $\widehat{I}(F_i;Y)$, $\widehat{I}(Y;C)$ using $\mathcal{A}_{t_i}$\\
   % \STATE Get a new feature $F_i$ at time $t_i$;
  %  \STATE\;{\bf for} each feature $Y\in S^*_{t_{i-1}}$ {\bf Compute} 
  %  \STATE \qquad  $\widetilde{I}^{i\ell}(F_i;C)$, using sample $\mathcal{X}_{i\ell}$ for $\ell=1,\ldots,m$ 
  %  \STATE \qquad Find $\mathcal{X}^*_{i}={\rm argmax}\limits_{\mathcal{X}_{i\ell}} \widetilde{I}^{i\ell}(F_i;C)$ and let $\widetilde{I}^{i*}$ be MI estimator using sample $\mathcal{X}^*_i$;
   \STATE \qquad{\bf if} $\widetilde{I}(F_i;C)\leq \delta$ {\bf then}
   \STATE \qquad\quad Discard $F_i$ and go to Step 19
   \STATE \qquad{\bf end}
   \STATE \qquad{\bf else}
 %   \STATE \qquad{\bf if} $I(F_i,X|C)\leq \delta$ {\bf then}
     \STATE \qquad\quad{\bf for} each feature $Y\in S^*_{t_{i-1}}${\bf do}
     \STATE \qquad\quad\quad{\bf if} $\widetilde{I}(Y;C)> \widetilde{I}(F_i;C)$  and $\widetilde{I}(F_i;Y)\geq \widetilde{I}(F_i;C)$ {\bf then}
      \STATE \qquad\quad\quad Discard $F_i$ and go to Step 19
   \STATE \qquad\quad\quad{\bf end}
     \STATE \qquad\quad\quad{\bf if} $\widetilde{I}(Y;C) < \widetilde{I}(F_i;C)$ and $\widetilde{I}(F_i;Y)\geq \widetilde{I}(Y;C)$ {\bf then}
     \STATE \qquad\quad\quad\;$S^*_{t_{i-1}}=S^*_{t_{i-1}}-Y$
      \STATE \qquad\quad\quad{\bf end}
   %  \STATE \qquad\quad\quad{\bf if} $\exists$ subset$\subseteq S^*_{t_{i-1}}-\{Y\}$ s.t. $G_{\rho}(Y;{\rm subset}|C)\geq \gamma$ {\bf then}
   %   \STATE \qquad\quad\quad\;$S^*_{t_{i-1}}=S^*_{t_{i-1}}-Y$
     %  \STATE \qquad\quad\quad{\bf end}
     \STATE \qquad\quad{\bf end}
     \STATE \qquad{\bf end}
     \STATE $S^*_{t_{i}}=S^*_{t_{i-1}}\cup F_i$
     \STATE {\bf until} no features and sample are streaming;
   \ENSURE  $S^*_{t_{i}}$: the selected feature set with fixed size $d$
  \end{algorithmic} 
  \caption{The SAOLA Algorithm with Streaming Samples}
  \label{algorithm1}
  \end{algorithm}

 The work closest in structure to our proposed GOA algorithm is SAOLA~\cite{Wuetal.2013}.
 Apart from using standard mutual information, this approach was designed with the assumption that the number of samples was fixed.
 To ensure a fair comparison to GOA under OSFS-SS we extend SAOLA to work with streaming samples, as shown in Alg.~\ref{algorithm1}.
 Given a stream of data $\mathcal{A}_{t_i}$, $\widehat{I}$ denotes a histogram estimator for standard MI.

 %\st{The most relevant algorithm to the proposed GOA algorithm is SAOLA \cite{Wuetal.2013}. In this approach standard mutual information is used to determine dependency. However the algorithm is designed for the streaming setting where sample is fixed. Therefore below we provide SAOLA with streaming sample. Given stream data $\mathcal{A}_{t_i}$ denote $\widehat{I}$ an histogram estimator of standard MI $I$.}
 The main distinction between the original and extended version of SAOLA is the selection of the data stream with the largest MI corresponding to the class variable, $C$.
  This method is key to filtering out the most informative stream of data among multiple streams corresponding to the same feature.
  Once the data block is selected, the extended SAOLA algorithm functions similarly to the original.

\end{document}